\documentclass[11pt,a4paper]{article}
\usepackage[a4paper]{geometry}
\RequirePackage{fix-cm}

\title{Matheuristics to  optimize  refueling and maintenance planning  of nuclear power plants}
\author{Nicolas Dupin, El-Ghazali Talbi\\
nicolas.dupin@universite-paris-saclay.fr}

\date{Accepted for publication in Journal of Heuristics, Springer, in July 2020}

\usepackage{amssymb}
\usepackage{graphicx}
\usepackage{amsmath}
\usepackage{url}
\def\N{{\mathcal{N}}}

\def\CC{{\mathcal{C}}}

\def\II{{\mathcal{I}}}
\def\JJ{{\mathcal{J}}}
\def\PP{{\mathcal{P}}}
\def\WW{{\mathcal{W}}}
\def\KK{{\mathcal{K}}}
\def\MM{{\mathcal{M}}}

\usepackage{color}

\begin{document}

\maketitle


\color{blue}{
This article  should be cited as
Dupin, N., Talbi, E. Matheuristics to optimize refueling and maintenance planning of nuclear power plants. J Heuristics (2020). \url{https://doi.org/10.1007/s10732-020-09450-0}
}

\color{black}

\begin{abstract}
 Planning the maintenance of nuclear power plants is a complex optimization problem, 
 involving a joint optimization of maintenance dates, fuel constraints and power production decisions.
 This paper investigates Mixed Integer Linear Programming (MILP) matheuristics for this problem,
 to tackle large size instances used in operations with a time scope of five years, and few restrictions with time window constraints for the latest maintenance operations.
Several constructive matheuristics 
and  a Variable Neighborhood Descent  local search  are designed. 
 The matheuristics are shown to be accurately effective for medium and large size instances.
The matheuristics give also results  on the design of MILP formulations and  neighborhoods 
for the problem.
Contributions for the operational applications are also discussed.
It is shown that the restriction of time windows, which was used to ease computations, induces large over-costs and that this restriction is not required anymore with the capabilities of matheuristics or local searches
to solve such size of instances.
Our matheuristics can be extended to a bi-objective optimization extension with stability costs, for the monthly re-optimization of the maintenance planning in the real-life application.
\end{abstract}

\vskip 0.3cm

\noindent{\textbf{Keywords} : Hybrid heuristics;
 Matheuristics ; Mixed Integer Programming ;
 Maintenance planning; Nuclear Power Plants ; Optimization in Energy}

\section{Introduction}

Planning maintenance of nuclear power plants is a crucial issue in energy management, with huge
financial stakes in France \cite{Ren93}.
Indeed, the part of nuclear production in the French power production is around $60-70\%$,
and the marginal costs of nuclear production are lower than the ones of other thermal power plants \cite{Kh07}.
At a first glance, the nuclear power plants should be in maintenance  when the demands in electricity are low, especially in summer.
Technical and operational constraints must be satisfied for the maintenance planning, it makes difficult to plan the maintenance for human experts. 
Operations research approaches were investigated
since the 70's  \cite{dopazo1975optimal}.
 With the continuous  progress of hardware and algorithms, especially in Mixed Integer Linear Programming (MILP),
 it became possible to tackle more complex and realistic problems in the decision support \cite{Fourcade}.
 The  sizes of the real-life instances has always been a challenging issue.
 Heuristics and decomposition techniques 
 were used for the short term Unit Commitment (UC) problem, and also for the maintenance planning optimization  of nuclear power plants \cite{Dub05,Kh07}.
Recent progress in MILP allows an efficient resolution
without decomposition for UC problems with additional real life constraints (\cite{ucp}).
Hybridizing heuristics using mathematical programming, known as matheuristics \cite{maniezzo2009matheuristics},
allows to solve realistic size of instances \cite{dupin2016matheuristics,dupin2017multi}.

A problem of  refueling and maintenance planning of nuclear power plants was specified by  the French Utility company (EDF)
in the 2010 ROADEF/EURO Challenge, a biennial competition of operations research \cite{roadef}.
The best results of the challenge were mainly obtained in \cite{beImproved,gardi} with local search approaches.
MILP formulations were also investigated, many  constraints 
being linear in the specification \cite{roadef}. 
 Matheuristics  were also used to tackle the large size instances of the problem, as investigated by \cite{jost,lusby,Roz12}. 
 The large size of the instances was a bottleneck for many approaches, 
especially on two instances, denoted B8 and B9, which
 can be seen as the most realistic instances for the real-life operations.
In this paper, we focus on a deterministic version of the problem given for the Challenge.
The difficulty of the instances B8 and B9 is explained and reproduced generating new instances
from all the instances of the Challenge.
The first motivation  is to  solve accurately
 the large instances which are relevant for real life applications.
The second motivation is to give  results related to the design of
MILP formulations  and heuristics for the problem.

This paper is organized as follows.
In Section 2, we describe precisely the problem with its industrial context.
In Section 3, we mention related works to appreciate our contributions. 
 In Section 4, we present a MILP compact formulation.
In Section 5, constructive matheuristics are designed using the previous MILP formulation. 
In Section 6, a local search matheuristics with MILP neighborhoods
provides local improvements once feasible solutions are known. 
 In Section 7,  the computational results are discussed. 
 In Section 8,  our contributions  are summarized,  discussing also future directions of research.
Appendix A discusses some industrial and numerical justifications
related to the model definition, and Appendix B gathers  tables of intermediate results.

  \begin{table}
\noindent{
\caption{Definitions and notations for the input parameters}\label{setROADEF} \label{notationsROADEF}

 \noindent{\textbf{Sets and indexes }}
 
\begin{tabular}{ll}
$w \in \WW = [\![1, W]\!]$ & Weekly time steps.\\
$j \in \JJ = [\![1, J]\!]$ & Flexible (Type 1, T1) power plants.\\
$i \in \II = [\![1, I]\!]$ & Nuclear power plants (Type 2, T2).\\
$k \in \KK = [\![0, K]\!]$ & Cycles related to T2 units, $k=0$  for initial conditions.\\ 
$c\in \CC$ & Scheduling constraints, type CT14 to CT21 in the specification \cite{roadef}.
\end{tabular}
}

 \vskip 0.3cm
 
 \noindent{\textbf{Temporal notations }}

\noindent{
\begin{tabular}{ll}
$\mathbf{D}_{w}$ & Power demands at time step $w$.\\
$\mathbf{F}_w$ & Conversion factor between power and fuel.\\
\end{tabular}
}

\vskip 0.3cm

\textbf{Notations for T1 units $j$}

\noindent{
\begin{tabular}{ll}
$\mathbf{C}_{j,w}^p$ & Production Costs  proportional to the generated power at $w$.\\
$\mathbf{\underline{P}}_{j,w}$ & Minimal power to generate at time step $w$. \\
$\mathbf{\overline{P}}_{j,w}$ & Maximal generated  power at time step $w$. 
\end{tabular}
}

\vskip 0.3cm

\textbf{Notations for T2 units $i$}

\begin{tabular}{ll}
$\mathbf{\overline{A}}_{i,k}$ & Maximal fuel level remaining in cycle $k$ to process outage  $k+1$.\\
$\mathbf{Bo}_{i,k}$ & Fuel level "Bore Null" of cycle $k$\\
$\mathbf{C}^r_{i,k}$  & Proportional cost to the refueling levels in cycle $k$.\\
$\mathbf{C}^f_i$  & Proportional cost to the final remaining fuel levels at $W$.\\
$\mathbf{C}^{pen}_{i,k,w}$ & Additional (stability) cost to schedule outage $k$ at week $w$.\\
$W_{i,k}^0$ & Initial week where outage $k$ was scheduled\\
$\mathbf{Da}_{i,k}$ & Outage duration for maintenance and refueling at cycle $k$.\\
$\mathbf{{To}}_{i,k}$ & First possible outage week for cycle $k$ of T2 plant $i$.\\
$\mathbf{{Ta}}_{i,k}$ & Last possible beginning  week for outage $k$ of T2 plant $i$.\\
$\mathbf{\overline{P}}_{i,w}$ & Maximal generated  power at time step $w$ .\\
$\mathbf{Q}_{i,k}$ & Proportion of fuel that is kept during reload in cycle $k$ at plant $i$\\
$\mathbf{\underline{R}}_{i,k}$ & Minimal refueling at outage $k$.\\
$\mathbf{\overline{R}}_{i,k}$ & Maximal refueling at outage $k$.\\
$\mathbf{\overline{S}}_{i,k}$ & Maximal fuel level  of T2 plant $i$ at   production  cycle $k$.\\
$ \mathbf{Xi}_{i}$ & Initial fuel  stock of T2 plant $i$.
\end{tabular}

\vskip 0.3cm

\textbf{Notations for scheduling constraints $c \in \CC$}

\noindent{
\begin{tabular}{ll}
$A_c$ & Subset of outages involved in constraint $c$.\\
$W_c$ & Subset of time periods involved in constraint $c$.\\
$\mathbf{R}_{c,a,w}$ & Resource consumption  at time step $w$. \\
$\overline{R}_{c,w}$& Maximal global resource consumption at $w$. 
\end{tabular}
}

 \end{table}

\section{Problem statement}

This section describes the problem of refueling and  maintenance planning, its constraints and objectives
and some insights of the data characteristics and implementation constraints which induced our scientific problematic.
The differences with the 2010 ROADEF/EURO Challenge are also highlighted.

\subsection{{Simplifications from the 2010 ROADEF/EURO Challenge}} \label{simplSec}

We consider in this paper the following simplifications of the constraint of the  2010 ROADEF/EURO Challenge \cite{roadef}:

\begin{itemize}
 \item \emph{weekly time steps}:
 in the Challenge, there were two types of time steps, a weekly discretization for the outage dates
and a hourly or daily discretization for power production and demands. 
In this study, we consider only weekly time steps, production and demand levels are aggregated weekly.
 
 \item \emph{determistic model}: In the Challenge, discrete  scenarios modeled the uncertainty of  power demands, production capacities and costs. In this study, we consider only one deterministic scenario in the optimization.
 
 \item \emph{relaxation of CT6 constraint}: CT6 constraints express  deterministic production curves for nuclear power plants when the fuel is 
 lower than a known threshold, as illustrated Figure \ref{prodRoadef}.
 The relaxation of these constraints is justified numerically in Appendix A.
 
 \item \emph{relaxation of CT12 constraint}: CT12 constraints express an aggregated constraint,
 limiting the modulation possibilities of nuclear power plants, to enforce the nuclear power plants
 to produce mostly at their maximal level.
  The relaxation of these constraints is justified  in Appendix A, mentioning  realistic constraints
  related to modulation of nuclear power plants.
 
\end{itemize}


\subsection{{Set and indexation}} 
Table \ref{notationsROADEF} gathers the notation for the sets and indexation.
Two kinds of power plants are modeled. 
On one hand, Type-2 (or T2) power plants indexed with $i \in \II$, correspond to nuclear power plants.
T2 power plants have to be shut down for refueling and maintenance regularly.
On the other hand, Type-1 (or T1) power plants are indexed with $j \in \JJ$, model other power plants. 
\textit{Outages} and \textit{production campaigns} are indexed with the cycles $k \in \KK$ for all T2 plants $i$. 
By convention, a cycle $k$ begins with the outage  period for maintenance and refueling, before the { production campaign} $k$. 
The outage and production decisions are discretized weekly and indexed with $w \in \WW=[\![1;W]\!]$.
Lastly, $c \in \CC$ gather scheduling constraints in outages. Such constraints were detailed in the sets from CT14 to CT21 in the Challenge specification \cite{roadef}. Actually, 
one can model  such constraints in one common framework, as mentioned in Table \ref{contraintesROADEF}.

\begin{table}
\caption{Description of the constraints considered in this paper,
with the reference to the corresponding constraints from the EURO/ROADEF Challenge}\label{contraintesROADEF}
 \vskip 0.2cm
\begin{tabular}{l}
\hline
 \begin{minipage}{0.99 \linewidth}
 \vskip 0.2cm
 \textbf{CT1, demand covering}:  for all time step $w\in \WW$, 
 the total production of T1 and T2 power plants must equalize the demands $\mathbf{D}_{w}$. 
   \vskip 0.3cm
 \end{minipage}\\
 \hline
  \begin{minipage}{0.99\linewidth}
 \vskip 0.2cm
 \textbf{CT2, T1 production bounds} for all $w \in \WW$,
 the production domain of T1 plant $j\in \JJ$ describes exactly the continuous domain   $[\mathbf{\underline{P}}_{j,w}, \mathbf{\overline{P}}_{j,w}]$ 
 \vskip 0.3cm
 \end{minipage}\\
 \hline
%
 \begin{minipage}{0.99\linewidth}
 \vskip 0.2cm
  \textbf{CT3-5,  T2 production bounds}
The productions of  offline T2 power plants are null during outages.
Otherwise,  the production domain of T1 plant $j\in \JJ$ describes exactly the continuous domain   $[0,\mathbf{\overline{P}}_{i,w}]$
for all period $w\in \WW$.

     \vskip 0.3cm
 \end{minipage}\\
\hline

  \begin{minipage}{0.99\linewidth}
 \vskip 0.2cm
 \textbf{CT7, refueling quantities}:  
 The refueling possibilities for outage $k$ of T2 plant $i$ describes exactly the continuous domain $[\mathbf{\underline{R}}_{i,k},\mathbf{\overline{R}}_{i,k}]$
 \vskip 0.3cm
 \end{minipage}  
  \\
\hline
    \begin{minipage}{0.99\linewidth}
 \vskip 0.2cm
  \textbf{CT8, initial fuel stock}: The initial fuel stock for T2 unit $i$ is $\mathbf{Xi_i}$. 
 \vskip 0.3cm
 \end{minipage}\\
\hline

      \begin{minipage}{0.99\linewidth}
 \vskip 0.2cm
  \textbf{CT9, fuel consumption}: The
  fuel consumption for T2 unit $i$ at period $w$ 
  is  proportional to the generated power, with a proportional factor $- \mathbf{F}_w$.
   \vskip 0.3cm
 \end{minipage}\\
\hline

\begin{minipage}{0.99\linewidth}
 \vskip 0.2cm
  \textbf{CT10, losses in refueling}
  After an outage, the new fuel stock is the sum of the refueling quantities (cf CT7) and
  the amount of unspent fuel after the last production cycle.
  A proportional loss factor $\mathbf{Q}_{i,k}<1$ is applied to the residual fuel before refueling.
   \vskip 0.3cm
 \end{minipage}\\
\hline 
    
   \begin{minipage}{0.99\linewidth}
 \vskip 0.2cm
 \textbf{CT11, bounds on fuel stock} The fuel level is bounded in $[0,\mathbf{S}_{i,k}]$ for cycle $k$ of T2 unit $i$.
   The fuel level  must be lower than  $\mathbf{A}_{i,k+1}$ to process outage $k+1$. 
 \vskip 0.3cm
 \end{minipage}\\
 \hline
  \begin{minipage}{0.99\linewidth}
 \vskip 0.2cm
  \textbf{CT13, time windows}  The maintenance cycles follow the set $k\in \KK$ without skipping cycles: if outage $k+1$ is processed, it must follow the cycle $k$.
 The beginning dates of outage $k$ of T2 unit $i$
  must be included in $[\mathbf{To}_{i,k},\mathbf{Ta}_{i,k}]$. 
 \vskip 0.3cm

 \end{minipage}
  \\
\hline
  \begin{minipage}{0.99\linewidth}
 \vskip 0.2cm
  \textbf{CT14-21, resource scheduling constraints }: 
     For each constraint $c\in \CC$,  a subset of outages $A_c$  and a subset of time periods $W_c$   is considered.     For all outage  $a \in A_c$ and for all time period $w \in W_c$, a resource consumption $R_{c,a,w}$ is defined    for the outage $a$ being scheduled at $w$, and the global resource consumption at $w$ must be lower than $\overline{R}_{c,w}$
 \vskip 0.3cm
 \end{minipage}\\
 \hline
  \end{tabular}
 \end{table}

\subsection{{Constraint description}}
Table \ref{contraintesROADEF} defines the constraints with their nomenclature from CT1 to CT21 in the  specification \cite{roadef}.
CT1 are the demand constraints to equalize productions and demands for each  time step and each scenario.
Constraints CT2 to CT6 and CT12 model production constraints: production bounds for T1 and T2 power plants
and specific technical constraints of nuclear power plants.
Constraints CT7 to CT11 model stock level constraints: bounds on stock levels and refueling and CT9 is the equation linking T2 production and fuel consumption.
The remaining constraints (CT13 to CT21 ) are specific to T2 plants outage scheduling.
CT13 defines time window constraints to process maintenance.
Constraints CT14 to CT21 can be unified in a common format of resource constraints applying for the maintenance decisions, as defined in Table \ref{contraintesROADEF}.
These constraints express minimal spacing/ maximal overlapping constraints among outages, and 
Minimum spacing constraint between coupling/decoupling dates, 
resource constraints for maintenance using specific tools or manpower skills
and limitations of simultaneous outages with  maximal number of simultaneous outages or maximal T2 power off-line.

\subsection{{Objective function}}
The objective function  minimizes the power generation cost while satisfying the customer load for all time steps.
Production costs of T1 units $j$ are  proportional to the production levels. 
Production costs of T2 units are calculated proportionally to the fuel consumption:
 refueling costs  are considered, and reduced with
the  costs of the remaining fuel with a proportional factor $\mathbf{C}_{i}^{f}$ to avoid end-of-side effects.
This makes an aggregated cost function to optimize the global financial cost.

In  operations, the  maintenance planning  is re-optimized  monthly,
the baseline planning impacts the re-optimized one.
The  maintenance operations are outsourced to several external companies fixing the periods of operations.
Time window constraints CT13 can model such decisions
with hard constraints.
Actually, the maintenance planning can be slightly modified if the financial stakes worth the reorganization efforts.
Hence, we consider an objective of stability, defining with $\mathbf{C}^{pen}_{i,k,w}$ 
the stability cost to reschedule the maintenance  $(i,k)$ at week $w$.
Denoting  $W_{i,k}^0$ the initial beginning week of maintenance $(i,k)$,  $\mathbf{C}^{pen}_{i,k,W_{i,k}^0} = 0$. 
Any penalization with  $\mathbf{C}^{pen}_{i,k,W_{i,k}^0} = 0$ can be considered, like 
a linear penalization   $\mathbf{C}^{pen}_{i,k,w} = |w-W_{i,k}^0|$,
 a quadratic one $\mathbf{C}^{pen}_{i,k,w} = (w-W_{i,k}^0)^2$, 
or constant costs for modifications with 
$\mathbf{C}^{pen}_{i,k,w} = 1$ for all $w \neq W_{i,k}^0$.

\subsection{Data characteristics}

The datasets provided  for the Challenge are now public.
The instance characteristics are provided in Table \ref{instancesROADEF}.
Dataset A contains five small instances given for the qualification phase, in a horizon of five years, with 10 to 30 nuclear power plants having $6$ production cycles. 
Instances B and X  are more representative of real-world size instances, 
with 20 to 30 T1 units and around 50 T2 units.
These datasets are now public, dataset X was secret for the challenge, with instance characteristics that had to be symmetric to dataset B.
This symmetry holds for the cardinal of the different sets ,
but this does not hold for the number of binaries (i.e. the total amplitude of time windows). 
Table \ref{instancesROADEF} shows in the column \verb!varBin! the number of binary variables.
Instances B8 and B9 are more difficult. 
X13 and X14 are not symmetric with B8 and B9 in the number of binaries.

There is no time window constraints for cycles $k\geqslant 3$ for both instances B8 and B9,
which is not the case for the others.
Actually, B8 and B9 are the most relevant instances for the operational application.
Maintenance operations are outsourced with several companies which makes difficult to reorganize
maintenance dates in the first years of the planning horizon, whereas few such constraints apply 
for the last cycles of the planning horizon.
A special interest is given in our study to solve efficiently and accurately instances like B7, B8 and B9.

\begin{table}
\centering
\caption{Characteristics of the ROADEF  instances :
I,J number of T2 and T1 units, K number of cycles for all T2 unit, S number of scenarios, T,W,
number production and weekly time steps, varBin denotes the number of binaries}\label{instancesROADEF}
\begin{tabular}{|l|cc|cc|cc|c|}
\hline
Instances&I&J &K&S&T&W&varBin\\
\hline
A1&10&11&6&10&1750&250&3892\\
A2&18&21&6&20&1750&250&7889\\
A3&18&21&6&20&1750&250&8162\\
A4&30&31&6&30&1750&250&17465\\
A5&28&31&6&30&1750&250&15357\\
\hline
B6&50&25&6&50&5817&277&24563\\
B7&48&27&6&50&5565&265&35768\\
B8&56&19&6&121&5817&277&\bf{69653}\\
B9&56&19&6&121&5817&277&\bf{69306}\\
B10&56&19&6&121&5565&265&29948\\
\hline
X11&50&25&6&50&5817&277&20081\\
X12&48&27&6&50&5523&263&27111\\
X13&56&19&6&121&5817&277&30154\\
X14&56&19&6&121&5817&277&30691\\
X15&56&19&6&121&5523&263&27233\\
\hline
\end{tabular}
\end{table}
{}

\subsection{Hardware and software limitations}

For the Challenge, the hardware and software for the evaluation were specified.
The computer was  a bi-processor Intel Xeon 5420 2.5Ghz quad core (8GB of memory under Linux, 3GB under windows, 12MB of
cache). Cplex could be used for MILP solving in the version 12.1.2.0 for the evaluation.
Such hardware limitations are relevant for the industrial application, the software limitations may be updated to the current versions.
The time limit was set to one hour per instance for the evaluation of the Challenge.
This constraint was justified to be able to evaluate the approaches of many teams in the defined period (one month) between the
deadline for submitting the final program and the announcement of the challenge results.
For the real-life application, the planning is re-optimized monthly,
several hours for computing time are reasonable, and possibly computations by night \cite{dupin2015modelisation,griset2018methodes}.

\section{Related work}

This section  describes the state of the art of the problem.
The approaches of the 2010 EURO/ROADEF Challenge are of special interest,
focusing on the related works both in the heuristic and mathematical programming approaches.

\subsection{{General facts of the Challenge ROADEF}}

The approaches in competition  led to several publications of full papers
(\cite{Ang12,Gav13,Godskesen,gardi,jost,lusby,Roz12}). 
Slides   (mainly presented at the EURO 2011 conference) are also available at 
\url{http://www.roadef.org/challenge/2010/fr/methodes.php}.
We distinguish three main types of approaches: 
matheuristics based on MILP exact approaches, decomposition heuristics following the 2-stage structure
and straightforward local search approaches. These three types of approaches are analyzed precisely in the next section.

 A major difficulty came with the large size of instances.
Many approaches used  pre-processing reductions:
tightening the time windows with exact or heuristic pre-processing;  aggregating production time steps to week or aggregating the stochastic scenarios to the average one.
The heuristic pre-processing requires to repair the solutions obtained in a post-processing 
considering the whole problem. Reparation strategies  were in general effective to recover feasible solutions.
The instances B8-B9 raised the most difficulties among the competing approaches , with many approaches leading to large over-costs for these instances
whereas other instances have better results (we refer to the official results: \url{http://www.roadef.org/challenge/2010/en/results.php}).

\subsection{{Matheuristics based on MILP exact approaches}}
Two matheuristic approaches were designed for the  Challenge.
These matheurstics solve a simplified MILP problem before repairing the solution 
to ensure the feasibility for all the constraints and computing the cost in
the full model.
In both cases, the MILP model is solved heuristically using truncated exact methods,
to solve the instances in one hour.

Lusby et al. (2013) relaxed  the constraints CT6 and CT12, leading to a MILP formulation with binaries only for the maintenance decisions \cite{lusby}.
The production time steps were aggregated  weekly.
The stochastic scenarios were not aggregated, leading to 2-stage stochastic programming structure  solved by 
Bender's decomposition. 
The master problem concerns the maintenance decisions and the refueling quantities, 
 independent sub-problems are defined for each stochastic scenario 
with continuous variables for power productions and fuel levels.
The  heuristic of \cite{lusby} computes first the Linear Programming (LP) relaxation exactly with the Bender's decomposition algorithm.
Then, a Branch\&Bound (B\&B) phase repairs integrity, branching on binaries without adding new Bender's cuts.
Actually, a pool of solution is generated
to have several solutions to repair with respect to the constraints CT6-CT12.
Using a pool of solution  improves  feasibility issues and provides  better solutions than using only the best solution. 
The resulting heuristic approach was efficient for the small dataset of the qualification, difficulties occur for the final instances of the competition, especially B8 and B9 instances.

Rozenkopf et al. (2013) (\cite{Roz12}) considered an exact formulation of CT6 constraints,
in a Column Generation (CG) approach dualizing coupling constraints among units, i.e. CT1 demands and CT14 to CT21 scheduling constraints.
The MILP  considered a unique scenario, the average one, with production time steps aggregated weekly.
The CG approach  computes a LP relaxation relaxing scheduling constraints CT14-CT21,
with sub-problems solved by dynamic programming.
The CG heuristic adds these scheduling constraints in the ILP with the columns generated.
The ILP solution furnishes the final maintenance dates and week where the CT6 constraints are activated,
the cost and feasibility issues for the whole problem are given computing the production
by Linear Programming (LP).
This approach was one of the most effective for the challenge. 

We note that a  compact MILP formulation 
for the full problem was provided in \cite{Jonc10}, introducing binaries for all time steps, cycle and production mode $m \in \MM$.
This size of problem is not reasonable for a B\&B search, even in a truncated mode, this work was a preliminary
work before a CG work  similar with \cite{Roz12}.

\subsection{{Heuristics based on the 2-stage decomposition}}

A very common idea  was to decompose the problem in a 2-stage structure,
 distinguishing the optimization of maintenance and refueling of T2 units
 from the power production problems, as in \cite{Ang12,brandt,Bra13,Gav13,Godskesen,jost}.
 The first-level problem fixes the maintenance dates and the refueling levels modifying  the current solution while
 respecting  constraints CT7-CT11 and CT13-CT21.
 The current planning is  modified in a  neighborhood search, where the feasibility of constraints CT13-CT21 can be
 guaranteed using MILP of Constraint Programming models.
 The second-level sub-problems compute independently for each scenario the production plans
 optimizing the production costs and fulfilling the constraints CT1-CT6 and CT12
having fuel levels and maintenance dates fixed.
Such production problems can be solved using greedy strategies following increasing production costs
or using LP problems. 
Such decomposition approaches in competition were not among the most efficient.
 We note that the operational approach in EDF was in this scope before the Challenge,
we refer to \cite{Kh07}.

\subsection{{Local search approaches}}\label{LSneighborhood}

{Local search approaches}, tackling the problem  without decomposition, were used by
 the two best approaches of the Challenge, having  similar results,
on one hand a methodology similar with LocalSolver \cite{gardi,benoist2011localsolver}
and on the other hand an unpublished Simulated Annealing local search \cite{beImproved}.
We note that excellent results are reported in \cite{gardi} with a time limit of few minutes, which is interesting for an industrial application.
The choices of the neighborhoods is crucial for such approaches to have efficient moves
and cost computation of the  locally modified solution,
to be able to explore aggressively a lot of solutions.
The  neighborhoods chosen by \cite{gardi} are reported in their paper:


\begin{itemize}
 \item k-MoveOutagesRandom: select k outages among T2 plants randomly and move them randomly in “feasible” time intervals,  ensuring the respect of earliest and latest starting dates (CT13 constraints) and maximum stocks before and after refueling (CT11 constraints);
 
 \item k-MoveOutagesConstrained: select T2 plants which are involved in scheduling constraints CT14-CT21 randomly, select k outages of these plants randomly, move these outages randomly in feasible time intervals;

 \item k-MoveOutagesConsecutive: select a T2 plant randomly, select k consecutive outages of these plant randomly, move these outages randomly in feasible time intervals.
\end{itemize}

One can give explanations to the better quality solutions of  local search approaches.
Many exact optimization  and decomposition heuristics 
restricted the model with heuristic pre-processing, it induce over-costs, which also explains the better results of \cite{gardi,beImproved} that avoided such over-costs.
Furthermore, heuristic decomposition and  exact approaches are slowed down by the use of MILP, LP or CP computations.
Local search approaches explore aggressively more solutions.
Furthermore, the neighborhoods used for decomposition heuristics were usually small (e.g allowing to move only one an outage from only one week).
Such neighborhoods were used by \cite{beImproved} in the qualification phase, and led to bad solutions.
The effort to  have diversified and large neighborhoods, as mentioned in \cite{gardi}, is crucial.

\subsection{{Other works after the Challenge}} 
 
After the Challenge, other  issues were investigated.
One open  question was to compute efficiently dual bounds. 
The first dual bounds were furnished by\cite{Bra13} using dual heuristics,
and were improved in \cite{dupin2016dual,dupin2018dual}. 
Semi-definite relaxations were investigated, to improve bounds obtained with LP relaxations,
but the size of the instances 
is a bottleneck for such approaches \cite{Gor12}.
Using the optimal control theory, one can
consider the fuel and production constraints as a discretized Partial Derivative Equations system. 
It raised new perspectives  for local search approaches:
a sensitivity analysis can indicate moving directions, as in   \cite{barty2014sensitivity}.
However, the method do not allow moves that swap maintenance orders, which is interesting among
the nuclear reactors from the same site with constraints CT14, as used by \cite{gardi}.
Mathematical Programming methodologies were investigated to complexify the model,
searching for a robust scheduling considering delays in the maintenance operations.
Such extension were studied with CG approaches \cite{pira2014column,griset2018methodes},
and also with  multi-objective matheuristics \cite{dupin2016meta,griset2018methodes}.
For these last studies, computations are tractable restricting computations to horizons of 2-3 years.

\section{MILP formulation}

In this section, we provide a MILP formulation for the problem, relaxing only constraints CT6 and CT12 similarly to \cite{lusby}.
It leads to a MILP formulation where the only binary variables are the decisions of outage beginning weeks. 

\begin{table}[ht]
      \centering
\caption{Set of variables}\label{setVar}
      
\begin{tabular}{ll}
$d_{i,k,w}\in \{0,1\}$ & Beginning dates of outages of unit $i$ at cycle $k$ at $w$.\\
$r_{i,k}\geqslant 0$ & Refueling levels of unit $i$ at cycle $k$ . \\
$p_{i,k,w}\geqslant 0$ & Production of T2 unit $i$ at cycle $k$ at $w$. \\
$p_{j,w}\geqslant 0$ & T1 production of unit $j$ at week $w$ . \\
$r_{i,k}\geqslant 0$ & Refueling levels for unit $i$ at cycle $k$. \\
$x_{i}^{f}\geqslant 0$ & Fuel stock of T2 unit $i$ at the end of the optimizing horizon. \\
$x_{i,k}^{i}\geqslant 0$ & Fuel levels of T2 unit $i$ at the beginning of cycle $k$. \\
$x_{i,k}^{f}\geqslant 0$ & Fuel levels of T2 unit $i$ at the end of production cycle $k$.
\end{tabular}
\end{table}

\subsection{Definition of  variables}

The definition of variables is highlighted in Table \ref{setVar}.
We define variables $d_{i,k,w}$ for maintenance dates for each cycle $i,k$ with
  $d_{i,k,w}=1$  if and only if the outage $i,k$  begins before week $w$.
These are the only integer  variables thanks to the relaxation of CT6 and CT12, as in \cite{lusby}.


  
Other continuous variables are introduced: refueling quantities $r_{i,k}$ for each outage $(i,k)$,
T2 power productions $p_{i,k,w}$ at cycle $k$, fuel stocks at the beginning of campaign $(i,k)$ (resp at the end)  $x_{i,k}^{i},x_{i,k}^{f}$,
T1 power productions $p_{j,w}$, and fuel stock $x_{i}^{f} $ at the end of the optimizing horizon.
We note that T2 power productions $p_{i,k,w}$ are duplicated for all cycle $k$ to have a linear model, 
$p_{i,k,w}=0$ if week $w$ is not included in the production cycle $k$.

%

\subsection{{MILP formulation }}
The variable definition induces the MILP formulation below.
\begin{eqnarray}
 \displaystyle v_0 = \min_{d \in \{0,1\}^N, r,p,x \geqslant 0} & \displaystyle\sum_{i,k} \mathbf{C}^{r}_{i,k} r_{i,k} +  
\sum_{j,w} \mathbf{C}^{p}_{j,w} \mathbf{F}_w \: p_{j,w} - \sum_{i}  \mathbf{C}_{i}^{f} x_{i}^{f}
 & \label{PANobj} \\
\forall i , k, w, & d_{i,k,w-1}\leqslant d_{i,k,w} & \label{PANprecedence}\\
\forall i , k, & d_{i,k,\mathbf{To}_{i,k}-1}\leqslant 0 & \label{PANtw0} \\
\forall i , k, & d_{i,k,\mathbf{Ta}_{i,k}}\geqslant 1 & \label{PANtw1}\\
\forall w, & \sum_{i,k} p_{i,k,w} + \sum_j p_{j,w} = \mathbf{D}_{w}\label{PANdemand}\\
\forall  j,w, & \mathbf{\underline{P}}_{j,t} \leqslant p_{j,w} \leqslant \mathbf{\overline{P}}_{j,t}\label{PANflexPower}\\
\forall i,k,w,  &  p_{i,k,w} \leqslant \mathbf{\overline{P}}_{i,w} (d_{i,k,w -{\mathbf{Da}_{i,k}}} - d_{i,k+1,w} )\label{PANcoupling}\\
\forall i,k, & \mathbf{\underline{R}}_{i,k} \; d_{i,k,W} \leqslant r_{i,k} \leqslant \mathbf{\overline{R}}_{i,k}\; d_{i,k,W}\label{PANrefuel}\\ 
\forall i, & x_{i,0}^{i} = \mathbf{Xi}_{i}\label{PANfuelInit}\\
\forall i,k, & x_{i,k}^{f} = x_{i,k}^{i} - \sum_w \mathbf{F}_w \:  p_{i,k,w} \label{PANconso}\\
\forall i,k,  &  x_{i,k}^{i} - \mathbf{Bo}_{i,k} = r_{i,k} + \frac{\mathbf{Q}_{i,k} -1}{\mathbf{Q}_{i,k}} (x_{i,k-1}^{f} - \mathbf{Bo}_{i,k-1})\label{PANpertes}\\
\forall i,k, &   x_{i,k}^{i}   \leqslant \mathbf{S}_{i,k} \label{PANmaxStock}\\
\forall i,k, & x_{i,k}^{f} \leqslant \mathbf{A}_{i,k+1} + (\mathbf{S}_{i,k} - \mathbf{A}_{i,k+1})  (1-d_{i,k+1,W}) \label{PANanticip}\\
\forall i,k,  & x_{i}^{f} \leqslant x_{i,k}^{f} + \overline{S}_{i}    (1-d_{i,k,W} + d_{i,k+1,W} )\label{PANfuelFinal}\\
  \forall c\in \CC,w\in W_c, & \sum_{(i,k)\in { \mathbf{A}}^c} ({R}_{c,a,w} - {R}_{c,a,w-1}) d_{i,k,w} \leqslant \overline{R}_{c,w} \label{PANordo}
\end{eqnarray}

(\ref{PANprecedence}) is required with definition of variables $d$.
(\ref{PANtw0}) and (\ref{PANtw1}) model CT13 time windows constraints: outage $(i,k)$ is operated between weeks $\mathbf{To}_{i,k}$ and $\mathbf{Ta}_{i,k}$.
(\ref{PANdemand}) models CT1 demand constraints.
(\ref{PANflexPower}) models CT2 bounds on T1 production.
(\ref{PANcoupling}) models CT3, CT4 and CT5 bounds on T2 production.
(\ref{PANrefuel}) models CT7 refueling bounds, with a null refueling when outage $i,k$ is not operated, ie $ d_{i,k,W}=0$.
(\ref{PANfuelInit}) writes CT8 initial fuel stock.
(\ref{PANconso}) writes CT9  fuel consumption constraints on stock variables of cycles $k$ $x_{i,k}^{i},x_{i,k}^{f}$.
(\ref{PANpertes}) models CT10  fuel losses at refueling.
(\ref{PANmaxStock}) writes CT11 bounds on fuel stock levels on  variables $x_{i,k}^{i}$ thanks to the decreasing law (\ref{PANconso}).
(\ref{PANanticip}) models CT11 min fuel stock before refueling, these constraints are active for a cycle $k$ only if the cycle is finished at the end of the optimizing
horizon, ie if $d_{i,k+1,W}=1$, which enforces to have disjonctive constraints where case $d_{i,k+1,W}=0$ implies a trivial constraints thanks to (\ref{PANmaxStock}).
(\ref{PANfuelFinal})  linearizes the constraints to enforce $x_{i}^{f}$ to be the fuel stock at the end of the time horizon.
$x_{i}^{f}$ is indeed the $ x_{i,k}^{f}$ such that $d_{i,k,W}=1$ and $d_{i,k+1,W}=0$, for the disjonctive constraints (\ref{PANfuelFinal}) 
that write a trivial constraints in the other cases thanks to (\ref{PANmaxStock}), we define $\overline{S}_{i}= \max_{k} \mathbf{S}_{i,k}$.
(\ref{PANordo}) is a common framework for scheduling constraints from CT14 to CT21, which was noticed independently in \cite{lusby,jost}.

\subsection{Adding stability cost for dynamic re-optimization}

The objective function (\ref{PANobj})
considers only the financial cost similarly with \cite{roadef}.
With the extension of stability cost  defined in Section 2.2, we consider two objectives,
the financial cost $\mathbf{C}^{roadef}$ and the desorganization cost $\mathbf{C}^{desorg}$:

\begin{eqnarray}
 && \mathbf{C}^{roadef} = \displaystyle\sum_{i,k} \mathbf{C}^{r}_{i,k} r_{i,k} +  
\sum_{j,w} \mathbf{C}^{p}_{j,w} \mathbf{F}_w \: p_{j,w} - \sum_{i}  \mathbf{C}_{i}^{f} x_{i}^{f}
  \label{PANobjFin} \\
 && \mathbf{C}^{desorg} =  \displaystyle\sum_{i,k,w} \mathbf{C}^{pen}_{i,k,w}  (d_{i,k,w} - d_{i,k,w-1})  \label{PANobjStab}
\end{eqnarray}

The two objectives can be adversarial, the solution of the bi-objective optimization
is thus a set of Pareto non-dominated solutions.
We note that in the case  $\mathbf{C}^{pen}_{i,k,w}$ is defined by  $\mathbf{C}^{pen}_{i,k,w} = 1$ for all $w \neq W_{i,k}^0$.
and $\mathbf{C}^{pen}_{i,k,W_{i,k}^0} = 0$, i.e. $\mathbf{C}^{desorg}$ counts the number of desorganization.
This makes the objective $\mathbf{C}^{desorg}$ integer with a granularity of $1$, which is interesting considering an $\varepsilon$-constraint method with $\varepsilon =1$ (\cite{laumanns2006efficient}).
In this case, the constraint to bound the number of modifications of the original planning
by at most $\mathbf{Nmax}$ is given by:
\begin{equation}
 \sum_{i,k } (1 + d_{i,k,W^0(i,k)-1} - d_{i,k,W^0(i,k)}) \leqslant \mathbf{Nmax}
\end{equation}

\section{Constructive matheuristics}

 \subsection{Exact and heuristic pre-processing}

 As noticed during the Challenge, heuristic and exact pre-processing 
may reduce the size of the problem to improve the solving capabilities.
Exact pre-processing strategies fix  necessarily optimal decisions,
whereas heuristic  strategies reduce the problem with approximations inducing possibly over-costs.


Time window constraints CT13 can be tightened in a exact pre-processing. 
initial stocks defined in CT8, maximal T2 productions CT5, max stock before refueling with CT11.
This tightens by inference the time window constraints, as described in \cite{DellAmico,dupin2018dual,lusby}.
Some outages can also be removed when their earliest completion time exceed the time horizon.

There is no maximal length on a production cycle in our model and in  \cite{roadef}: 
decreasing profile phases can be followed by a nil production phase without any maximal duration in this state.
However, such situations are not economically profitable as the T2 production cost is lower than the T1 production cost. 
Thus, \cite{lusby} avoided such situations, defining heuristically maximal duration for production cycles and  tightening
by inference the latest dates to begin outages $\mathbf{{Ta}}_{i,k}$.
This is a heuristic pre-processing, the reduction of the solution space can be useful for exact methods. 
Similarly with \cite{dupin2017multi}, another heuristic pre-processing  fixes variables guided by the LP relaxation.
A LP-based heuristic reduction can be applied tightening the time window constraints with the integral values in the LP relaxation of variables $d_{i,k,w}$.

 \subsection{Post-processing to repair partial solutions}

 Another application of the stability extension is to repair an initial solution $W^1(i,k)$ of maintenance planning,
that may be infeasible, looking similarly with Feasibility Pump (FP, \cite{FeasPumpMIP}) for
the closest feasible solution 
using  $\mathbf{C}^{pen}_{i,k,w} = |w-W_{i,k}^1|$ or  $\mathbf{C}^{pen}_{i,k,w} = (w-W_{i,k}^1)^2$
and minimizing $\mathbf{C}^{desorg}$ or 
$\mathbf{C}^{desorg} + \epsilon \mathbf{C}^{roadef}$ with $\epsilon >0$ small.
Furthermore, one can also restrict the time window possibilities around the baseline solution
to avoid having a too large model

\subsection{{A  simplified MILP formulation}}
One can approximate the previous MILP formulation,
assuming that the T2 production is either null during outages, either
at its maximal level, ie
$ p_{i,k,w} = \mathbf{\overline{P}}_{i,w} (d_{i,k,w -{\mathbf{Da}_{i,k}}} - d_{i,k+1,w})$.
Imposing such constraints can lead to infeasibilities, it requires some other relaxations.
T2 power can be upper than some low demands $\mathbf{D}^{w}$, leading infeasibilities with \ref{PANdemand}. 
Relaxing \ref{PANdemand} with minimal production inequalities fix such problems.
Furthermore, a maximal T2 production can be infeasible with the fuel constraints.
More specifically, a maximal T2 production (and thus a maximal fuel consumption),
can be in contradiction with CT13 earliest dates of outages and positivity of fuel levels
To avoid such infeasibilities, we introduce continuous variables $\Delta_{i,k}$ 
to authorize negative fuel stocks in soft constraints with  high values $C_{i,k}^{\Delta}$ in the objective function.
It leads to following simplified MILP formulation:

\begin{eqnarray*}
v^{simpl} = \min & \sum_{i,k} \mathbf{C}^{r}_{i,k} r_{i,k} + 
\sum_{j,w}  \mathbf{C}^{p}_{j,w}  \mathbf{F}_w \: p_{j,w} - \sum_{i}   \mathbf{C}_i^{f} x_{i}^{f} + \sum_{i} C_{i,k}^{\Delta} \Delta_{i,k} & \\
 & A^{ordo} d \geqslant a^{ordo} & \\
 & \mathbf{\underline{R}}_{i,k} \; d_{i,k,W} \leqslant r_{i,k}   \leqslant \mathbf{\overline{R}}_{i,k}\; d_{i,k,W} & \\
\forall i,k, &   x_{i,k}^{i}   \leqslant \mathbf{\overline{S}}_{i,k}\\
\forall i,k, & x_{i,k}^{f} = x_{i,k}^{i} - \sum_w \mathbf{F}_w \:  p_{i,k,w} + \Delta_{i,k} \\
\forall i,k, & x_{i,k}^{f} \leqslant \mathbf{A}_{i,k+1} + (\mathbf{\overline{S}}_{i,k} - \mathbf{A}_{i,k+1}) \; (1-d_{i,k+1,W})\\
\forall i,k,  &  x_{i,k}^{i} - \mathbf{Bo}_{i,k} = r_{i,k} + \frac{\mathbf{Q}_{i,k} -1}{\mathbf{Q}_{i,k}} (x_{i,k-1}^{f} - \mathbf{Bo}_{i,k-1})\\
\forall i,k,  & x_{i}^{f} \leqslant x_{i,k}^{f} + \mathbf{\overline{S}}_{i}    (d_{i,k,W} - d_{i,k+1,W} )\\
\forall  j,w, & \mathbf{\underline{P}}_{j,w} \leqslant p_{j,w} \leqslant \mathbf{\overline{P}}_{j,w}\\
\forall i,w,  &  p_{i,k,w} = \mathbf{\overline{P}}_{i,w} (d_{i,k,w -{\mathbf{Da}_{i,k}}} - d_{i,k+1,w} )\\
\forall w, & \sum_{i,k} p_{i,k,w} + \sum_j p_{j,w} \geqslant \mathbf{D}_{w}\\
 & d \in \{0,1\}^N, r,p,x, \Delta \geqslant 0
\end{eqnarray*}

This MILP formulation  computes a first maintenance planning.
Then,  the induced production costs are computed as an LP, ensuring feasibility of all the constraints.
If this high level planning cannot lead to feasible production and fuel levels, a local reparation is required, as described in 5.2.

\subsection{{Restrict- Relax-and-Fix algorithm}}

Relax-and-Fix (RRF) algorithms (\cite{guzelsoy2013restrict}) construct iteratively a solution
following a partition of the variables, fixing values from the previous computations, and relaxing continuously others.
A RRF strategy can  partition the variables following the cycle index $k$.
It computes successively the solutions for cycles $k$, the dates of cycles $k'<k$
are fixed by the previous optimizations, and the variables related to  the cycles $k'>k$
are relaxed continuously or  simplified using a parametric aggregation  \cite{dupin2016dual}.

 \begin{figure}[ht]
 \centering 
\begin{tabular}{ l }
\hline
\textbf{Algorithm 1: Restrict- Relax-and-Fix algorithm}\\
\hline

 

\textbf{Initialisation:} define matrix $W_{i,k} = 0 $ for all $i \in \II, k \in \KK$.\\
 
\textbf{for} $k=1$ to $K-1$\\

\verb!     ! solve the MILP $v_{RRF}(k)$ with $\mathbf{Ta}_{i,k'} = \mathbf{To}_{i,k'} = W_{i,k'}$ for $k'<k$ \\

\verb!     ! store the solutions $d_{i,k}$\\

\verb!     ! update $W_{i,k} = \sum_w (1- d_{i,k}) $\\



\textbf{end for}\\

solve the MILP $v_0$ with $\mathbf{Ta}_{i,k'} = \mathbf{To}_{i,k'} = W_{i,k'}$ for $k'<K$ \\

\textbf{return} the  solution of the last MILP\\
\hline
\end{tabular}
\end{figure}

Feasibility difficulties  occur when a cycle optimization
fixes outages and no feasible solution exist to place the next outages because of  constraints CT13-CT21 and/or fuel constraints CT11.
 This induces in Algorithm 1 to consider binaries for cycle $k$ and $k+1$
for the iteration placing outages $k$, and to relax continuously/simplify formulations only for cycles $k'>k+1$.
The remaining fuel costs are relaxed also to focus on the local optimization around the cycles to optimize, aggregating
the cycles $k>k_0$ in the cycle $k_0+1$ relaxing the outage and the fuel constraints, as in \cite{dupin2016dual}.
The MILP formulation considered for each iteration is thus:


\begin{eqnarray}
v_{RRF}(k^0) =  \min & \displaystyle \sum_{i,k} \mathbf{C}^{rld}_{i,k} r_{i,k} + 
\sum_{j,w}   \mathbf{C}^{prd}_{j,w}  \mathbf{F}_w \: p_{j,w}& \\
\forall i , k\leqslant k_0+1, w, & d_{i,k,w-1}\leqslant d_{i,k,w} & \\
\forall i , k\leqslant k_0+1, & d_{i,k,\mathbf{To}_{i,k}-1}\leqslant 0\\
\forall i , k\leqslant k_0+1, & d_{i,k,\mathbf{Ta}_{i,k}}\geqslant 1 \\
\forall w, & \sum_{i,k\leqslant k_0+1} p_{i,k,w} + \sum_j p_{j,w} = \mathbf{D}^{w}\label{PANdemandRRF}\\
\forall  j,w, & \mathbf{\underline{P}}_{j,t} \leqslant p_{j,w} \leqslant \mathbf{\overline{P}}_{j,t}\label{PANflexPowerRRF}\\
\forall i,k\leqslant k_0+1,w,  &  p_{i,k,w} \leqslant \mathbf{\overline{P}}_{i,w} (d_{i,k,w -{\mathbf{Da}_{i,k}}} - d_{i,k+1,w} )\label{PANcouplingRRF}\\
\forall i,k\leqslant k_0+1, & \mathbf{\underline{R}}_{i,k} \; d_{i,k,W} \leqslant r_{i,k} \leqslant \mathbf{\overline{R}}_{i,k}\; d_{i,k,W}\label{PANrefuelRRF}\\ 
\forall i,k\leqslant k_0+1, & x_{i,k}^{f} = x_{i,k}^{i} - \sum_w \mathbf{F}_w \:  p_{i,k,w} \label{PANconsoRRF}\\
\forall i,k\leqslant k_0+1,  &  x_{i,k}^{i} - \mathbf{Bo}_{i,k} = r_{i,k} + \frac{\mathbf{Q}_{i,k} -1}{\mathbf{Q}_{i,k}} (x_{i,k-1}^{f} - \mathbf{Bo}_{i,k-1})\label{PANpertesRRF}\\
\forall i,k\leqslant k_0+1, &   x_{i,k}^{i}   \leqslant \mathbf{S}_{i,k} \label{PANmaxStockRRF}\\
\forall i,k\leqslant k_0, & x_{i,k}^{f} \leqslant \mathbf{A}_{i,k+1} + (\mathbf{S}_{i,k} - \mathbf{A}_{i,k+1})  (1-d_{i,k+1,W}) \label{PANanticipRRF}\\
\forall i,k\leqslant k_0+1,  & x_{i}^{f} \leqslant x_{i,k}^{f} + \overline{S}_{i}    (d_{i,k,W} - d_{i,k+1,W} )\label{PANfuelFinalRRF}\\
\forall c\in \CC,w\in W_c, & \sum_{(i,k\leqslant k_0+1)\in { \mathbf{A}}^c} ({R}_{c,a,w} - {R}_{c,a,w-1}) d_{i,k,w} \leqslant \overline{R}_{c,w} \label{PANordoRRF}
\end{eqnarray}

\subsection{Construct, Merge, Solve \& Adapt }

 Another decomposition iterates   unit by unit, fixing in the current planning of the computed sub-problems, declaring variables as integer for the current unit to optimize,
 and relaxing continuously the other units.
 Such heuristic has several  weaknesses.
Firstly, having one infeasibible sub-problem stops  the construction.
Secondly, the choice of the iterated unit order has some influence in the final solution.

\begin{figure}[ht]
 \centering 
\begin{tabular}{ l }
\hline
\textbf{Algorithm 2: Construct, Merge, Solve \& Adapt}\\
\hline
\textbf{Input:} a repairing operator $\mathcal{R}$\\

\textbf{Initialization:} define matrix $W_{i,k} = -1 $ for all $i \in \II, k \in \KK$.\\
 
\textbf{for each} T2 unit $i$:\\

\verb!     ! define new demands  $\mathbf{D}_{new}^{w} = \frac {\mathbf{\overline{P}}_{i,w}} {\sum_{i'}\mathbf{\overline{P}}_{i',w} } \mathbf{D}^{w}$\\

\verb!     ! solve the MILP $v_0$  with $\II = \{i\}$ and new demands  $\mathbf{D}_{new}^{w}$\\



\verb!     ! store the maintenance planning $W_{i,k} = \sum_w (1- d_{i,k}) $\\

\textbf{end for}\\

Merge the maintenance plannings $W_{i,k}$.\\

Compute as an LP  the related production and fuel decisions.\\

\textbf{if} the current planning is unfeasible:\\

\verb!     ! repair the current solution with  $\mathcal{R}$ \\

\textbf{end if}\\

\textbf{return} the last current solution \\
\hline
\end{tabular}
\end{figure}

In Algorithm 2, a Construct, Merge, Solve \& Adapt (CMSA, \cite{blum2016construct}) strategy computes independently  (for a possible parallel computation) planning for all units, as if  the other units were T1 power plants.
Merging these planning is likely  unfeasible for the scheduling constraints CT14-CT21,
a second phase repair the feasibility around this "ideal" solution.

\section{POPMUSIC-VND matheuristic}
Once a feasible solution is built with previous constructive matheuristics, 
this section improves the current solution in a local search procedure.

\subsection{{General algorithm}} 

We present here the  local search algorithm described in the Algorithm 3.
The neighborhoods are defined  using MILP definitions, similarly to \cite{adamo2017mip,dupin2016matheuristics}.
Many MILP neighborhoods can be designed, and large neighborhoods may be explored.
Algorithm 3 alternates the choice of the neighborhood , similarly with multi neighborhood search approaches.
 It induces that a local optimum for Algorithm 3 is a local optimum for all the considered neighborhoods.
 The stopping criterion could be a maximal time limit or a maximal number of iterations, or being in a local extremum for all neighborhoods.

MILP neighborhoods may be explored for small sub-problems with optimal B\&B solving, as in \cite{Larrain2017}.
In this work, MILP neighborhoods are defined with three characteristics for an efficient B\& B search in defined time limits.
Parametrization is empiric, for a good trade-off between solution quality and time spent in MILP solving.
The MILP neighborhoods are  defined as following:

\begin{itemize}
 \item The \textbf{restriction of search space}: Variable fixations or other extra constraints that the current solution satisfies to have an easier B\&B solving.
 \item a \textbf{B\&B stopping criterion}: it must defined  so that the B\&B search is efficient in a short solving time. 
 Parametrizing short time limits is a first step, additional parameters like stopping an absolute or relative threshold  between the best primal and dual bounds
 can avoid to  waste time  without improving significantly the best primal solutions.
 \item a \textbf{specific parametrization for MILP solving }: for an efficient B\&B search in the defined time limit.
  Small neighborhoods need few specific MILP parametrization. 
  For a better efficiency in short resolution time for difficult sub MILPs, one can
 emphasize the heuristic search , disable or limit cutting plane passes and MILP preprocessing (with parameters \verb!mipemphasis!,  \verb!cutpass!, \verb!presol! parameter (in the terminology of Cplex).
\end{itemize}

%

 The current solution is the primal solution given by the last B\&B resolution and it is also defined
as warmstart for the next B\&B resolution to improve the efficiency of B\&B primal heuristics, enabling RINS or Local Branching heuristics from the beginning.
This ensures that the solution given by the MILP resolution is at least as good as the current solution at each iteration.
This algorithm is thus a steepest descent algorithm, similarly to MILP-VND as in \cite{Larrain2017}.

 \begin{figure}[ht]
 \centering 
\begin{tabular}{ l }
\hline
\textbf{Algorithm 3: multi neighborhood descent with MILP neighborhoods}\\
\hline
\textbf{Input:} an initial solution, a set and order of neighborhoods to explore\\

\textbf{Initialisation:} \verb!currentSol! = initSolution, $\mathcal{N}=$initial neighborhood.\\
 
\textbf{while} the stopping criterion is not met\\

\verb!     ! define the MILP  with incumbent \verb!currentSol! and the neighborhood $\mathcal{N}$)\\

\verb!     ! define  \verb!currentSol! as warmstart\\

\verb!     ! \verb!currentSol! = solveMILP(MILP,timeLimit( $\mathcal{N}$))\\

\verb!     ! $\mathcal{N}=$ \verb!nextNeighborhood!$(\mathcal{N})$ \\

\textbf{end while}\\
\textbf{return} CurrentSolution \\
\hline
\end{tabular}
\end{figure}

\subsection{{MILP neighborhoods}} 
Multiple types of large and variable neighborhoods can be  defined. 
We denote with ${W}_{i,k}^o$ the beginning week of outage $(i,k)$ in the current incumbent solution.
Neighborhoods define strategies to fix variable $d_{i,k,w}$ around the solution defined by  ${W}_{i,k}^o$.
Generic neighborhoods are first derived from \cite{rins,LocalBranching}:

\begin{itemize}
\item $\N^{rins}$ : Similarly to RINS heuristic \cite{rins}, variables are fixed if they are a common integer value in the LP relaxation and in the current solution.
  \item $\N^{LB}_k$: for k an integer, k-Local Branching neighborhood allows only $k$ modifications to the incumbent, adding constraints as in  \cite{LocalBranching}:

  $$\displaystyle \sum_{i,k,w : w \geqslant {W}_{i,k}^o} d_{i,k,w} + \sum_{i,k,w : w < {W}_{i,k}^o} (1-d_{i,k,w})  \leqslant k$$
\end{itemize}


The multi-index structure allows to define neighborhoods fixing variables along a type of indexation:
\begin{itemize}
 \item $\N_{I}^{units}$ unit selection: only T2 units $i  \in I \subset \II$ are re-optimized.

   \item $\N_{k,k'}^{cycles}$ : variables relative cycles $k''$ with $k\leqslant k'' \leqslant k'$ are re-optimized.  
 
   \item $\N_{(a,b)}^{TW}$:  outages are re-optimized in time windows $[{W}_{i,k}^o - a k - b,{W}_{i,k}^o + ak +b]$.
 
\end{itemize}

 $\N_{(0,b)}^{TW}$ is a re-optimization with a fixed radius $b$ around the incumbent,
 $\N_{(0,1)}^{TW}$ is larger than neighborhood widely used for the Challenge, moving one outage of one week.
 The linear dependence with $a>0$ induces a "funnel" structure, to model that a move on the first outages
 can imply larger moves for the succeeding outages.
 The weakness of such neighborhoods is that it may not be possible to swap the order of outages constrained with scheduling constraints like CT14 with short neighborhoods.
 $\N_{k,k'}^{cycles}$ is a first answer to have complementary neighborhoods to swap the order of outages.
 $\N_{I}^{units}$ can be implemented for all single units, in this case, it authorizes large neighborhoods,
 and the  \emph{Winter Jump} neighborhoods: it may be interesting to change radically an outage scheduling, from an autumn period to a spring period, 
 to avoid winter periods when the substituted production  cost is higher.

\subsection{{Sequences of neighborhoods}}

A key point in the Algorithm 3 is the sequence of neighborhoods.
Applying Algorithm 3 with a single type of neighborhoods, it allows to analyze  
the quality and number of local extrema.
 In this case, different neighborhoods are compared starting with the same initial solutions.
 Usually, VND local searches increase the size of the neighborhoods when a local minimum is reached.
It can be implemented for Local Branching neighborhoods $\N^{LB}_k$ with $k$ increasing , as developed in \cite{vnsLocalBranching} or with neighborhoods $\N_{(0,k)}^{TW}$.
In our case study, long computation time are allowed in the operational process for the monthly optimization of the maintenance planning.
We do not seek to find the best solutions in short resolution time.
Having many types of neighborhoods, the stopping criterion is firstly to reach a local minimum for all the types of neighborhoods.
The choice of neighborhoods is nested, alternating deterministically the order of neighborhoods
and stopping the Algorithm 3 when no neighborhood improves the current best solution.

Using neighborhoods $\N_{I}^{units}$  or $\N_{k,k'}^{cycles}$, one can design different partitions of the binary variables.
A partition of units $P \subset \PP(\II)$ give rise to neighborhoods $\N_{p}^{units}$ for $p \in P$
partitioning the variables of the problem.
$P=\{ \{i\} \}_{i \in \II}$ is a first partition units by units, reoptimizing all the planning separately.
One can consider partitions with subsets of at least $k$ elements where $k$ is given.
In the applications,  partition occurs in sites with several nuclear reactors,
in subsets of the constraints CT14-CT15.
VND iterations can  successively  re-optimize along different partitions, similarly with 
POPMUSIC ({Partial optimization} meta-heuristic under special intensification conditions, see \cite{taillard2002popmusic}).
$\N_{k,k}^{cycles}$ neighborhoods  define also a partition, "orthogonal" to the unit re-optimization.

\section{Computational results and analyses}

The computational experiments were computed with a personal computer
having a processor Intel(R) Core(TM) i7-4790S CPU @ 3.20GHz using up to 8 threads,
 running Linux Ubuntu 18.04 with 8 GB of RAM memory.
MILP and LP were solved with  Cplex version 12.8.0.
We note that, contrary to the straightforward MILP solving with Cplex, the final  matheuristic is   even scalable  on lighter hardware, with also older versions of Cplex.

\subsection{{New instances}}

We used the dataset from the EURO/ROADEF 2010 challenge presented in Section 2.5 and Table \ref{instancesROADEF},  aggregating production time steps to weeks and the stochastic scenarios to a single scenario of demands and costs as in \cite{Roz12}.
Small instances were also generated to have comparisons  with known optimal values. 
Instances suffixed with \verb!_3_120! consider at least $3$ cycles in the $120$ first weeks of the ROADEF instances.
All these instances \verb!_3_120! are solvable in less than one hour, except  \verb!B9_3_120!,
as shown in Table \ref{CplexConv}.

Instances are generated from the dataset B and X from the Challenge,
respectively suffixed with \verb!Ext2! and \verb!Ext3!,
removing the time window for cycles $k > 2$ and $k > 3$ respectively.
These instances were generated 
to reproduce instances with a similar difficulty from B8 and B9.
Table \ref{instances2} compares  the number of binaries of these new instances
to the original ones.
Table \ref{instances2} shows that the number of binaries is increasing with  \verb!Ext2! and \verb!Ext3! datasets,
\verb!Ext2! reduces considerably the gaps between B8 - B9 and the other instances regarding the number of binaries.
This gap closing is also observable after Cplex solving, as shown in Table \ref{CplexROADEFandExt2}.

Table \ref{gapBKS}  compares the Best Known primal Solutions (BKS) of the extensions Ext2, Ext3 to the one of the original dataset
for the instance sets B and X, except for B8 and B9.
Table \ref{gapBKS} shows that the heuristic reduction of time windows
has an impact of $1\%$, which is very significant regarding the financial value, and also comparing
to the over-cost induces by different optimization methods for the Challenge.
The numerical results aim to provide an accurate resolution of the difficult instances, which are the most relevant for the industrial application, with a validation on $18$ small instances where the optimal values are known.

\begin{table}
\centering

\caption{Comparison of the number of binaries for the original instances of the ROADEF Challenge,
and the reduced and extended instances. The numbers of binaries are shown without pre-processing
and using the exact pre-processing from \cite{dupin2018dual}}\label{instances2}
{
\begin{tabular}{|l|cc|cc|cc|cc|}
\hline
&ROADEF&ROADEF&\_3\_120&\_3\_120&Ext3&Ext3&Ext2&Ext2\\
Instance &&+PP&&+PP&&+PP&&+PP\\
\hline
A1&3878&497&171&171&-&-&-&-\\
A2&7853&1139&522&345&-&-&-&-\\
A3&8124&1017&398&298&-&-&-&-\\
A4&17384&2115&956&554&-&-&-&-\\
A5&15293&3008&1314&1163&-&-&-&-\\
\hline
B6&24485&3697&801&801&43069&10747&56179&20488\\
B7&35586&7500&3373&1232&43390&12736&52639&21519\\
B8&69378&26014&10825&3178&70876&28683&72367&30362\\
B9&69026&29208&10808&3663&70549&31915&72042&33942\\
B10&29843&4533&934&934&46332&11571&60302&23010\\
\hline
X11&20015&3610&810&810&43015&11657&56149&22390\\
X12&27000&4837&1249&1014&40362&11237&51672&20837\\
X13&30019&5169&1994&1118&49822&13236&63406&23241\\
X14&30559&5824&1975&1165&49831&14630&63404&26079\\
X15&27140&4144&983&983&45966&12272&59840&23785\\
\hline
\end{tabular} 
}
\end{table}

\begin{table}
\centering

\caption{Cost values for the BKS of the medium and large instances,
the over-cost gap to the BKS of the ROADEF instances are computed for the Ext2 and Ext3 extensions
to show the impact of a heuristic reduction of time windows.}\label{gapBKS}
{
\begin{tabular}{|l|c|cc|cc|}
\hline
&ROADEF&Ext3&Ext3&Ext2&Ext2\\
&BKS&BKS&gap&BKS&gap\\
\hline
B6&76966,1M&76734,8M&0,30 \%&76407,1M&0,73 \%\\
B7&74232,0M&73739,0M&0,66 \%&73544,9M&0,93 \%\\
B10&69501,5M&69143,1M&0,52 \%&68834,6M&0,96 \%\\
\hline
X11&73017,8M&72023,1M&1,36 \%&71801,2M&1,67 \%\\
X12&70599,5M&70240,4M&0,51 \%&70255,8M&0,49 \%\\
X13&69227,1M&68352,4M&1,26 \%&68063,2M&1,68 \%\\
X14&68395,3M&67799,8M&0,87 \%&67496,5M&1,31 \%\\
X15&66028,5M&65638,8M&0,59 \%&65271,8M&1,15 \%\\
\hline
Total&567967,8M&563671,4M&0,76 \%&561675,1M&1,11 \%\\
\hline
\end{tabular}
}
\end{table}

\subsection{{Table results}}

To compare different approaches on an instance $i$, we computed for each instance $i$ the cost of the best primal solution known, denoted $\emph{BKS}(i)$,
  without restriction in the time limit.
 Lower bounds (LB) or upper  bounds (UB), denoted $v(i)$ on instance $i$, are compared with the gap indicator:
${\emph{gap}_i = \dfrac {\mid v(i) - \emph{BKS(i)} \mid} {\emph{BKS}(i)}}$.
 Over-cost gaps over $100\%$ are considered as non significant solutions, and considered as a failure. 

To show results on a group of instances, we provide statistics 
 using the following indicators:
\begin{itemize}

\item  {$\bar{t}$}, the  average time solving in seconds;

\item  {$\bar{g}$}, the  average gap;

\item  {$\sigma_{g} =   \sqrt{\frac 1 {n} \sum_{i=1}^n ( \emph{gap}_i - \bar{g})^2}$} to measure dispersion among $n$ instances;
 \item  \emph{Q1}, \emph{Q2}, \emph{Q3} denote the  quartiles of the gap sequence, Q2 is the median;
 
 \item  {$N_{x}$} represent the number of instances with a gap below $x \%$;
 
 \item  {$N_{F}$} represent the number of failures, i.e. the number of instances with a gap upper $100 \%$;
 
\end{itemize}

 \begin{table}
\centering
\caption{Comparison to the gaps to the BKS for different LB and UB at different stage of the B\&B solving:
LB of the LP relaxation, LB and UB at the root note after cutting plane passes and before branching,
and within computation times of 1h and 3h.
The LB with parametric relaxation of \cite{dupin2016dual} is also provided.
The results are separated in two clusters. The 18 instances where the optimal value is known (A1, A2, A3, A4 and *\_3\_120 instances except  B9\_3\_120, are separated from the other, more difficult.
}\label{CplexConv}
\vskip 0.3cm
Results for the instances where the optimal value is known.
\vskip 0.2cm
\begin{tabular}{|ll|c|c|ccc|ccc|}
\hline
&&$\bar{t}$&$\bar{g}$&Q1&Q2&Q3&$N_{0.01}$&$N_{0.05}$& $N_{1}$\\
\hline
LB&LP&0,7&1,74 \%&0,22 \%&1,09 \%&1,79 \%&1&3&7\\
\hline
LB&nod1&5&0,33 \%&0,02 \%&0,10 \%&0,24 \%&5&5&17\\
UB&nod1&5&0,14 \%&0,00 \%&0,03 \%&0,06 \%&7&13&17\\
\hline
LB&1h&217&0,01 \%&0,01 \%&0,01 \%&0,01 \%&18&18&18\\
UB&1h&217&0,00 \%&0,00 \%&0,00 \%&0,00 \%&18&18&18\\
\hline
LB&\cite{dupin2016dual}&&0,93 \%&0,11 \%&0,28 \%&1,37 \%&1&2&13\\
\hline
\end{tabular}
\vskip 0.3cm
Results for the other instances where the optimal value is not known.
\vskip 0.2cm
\begin{tabular}{|ll|c|c|ccc|ccc|}
\hline
&&$\bar{t}$&$\bar{g}$&Q1&Q2&Q3&$N_{0.01}$&$N_{0.05}$& $N_{1}$\\
\hline
LB&LP&91&8,83 \%&4,55 \%&9,66 \%&12,05 \%&0&0&0\\
\hline
LB&nod1&808&7,37 \%&3,05 \%&8,05 \%&10,94 \%&0&0&4\\
UB&nod1&808&-&90,93 \%& -&- &0&0&6\\
\hline
LB&1h&3600&6,72 \%&1,62 \%&7,58 \%&10,38 \%&0&0&6\\
UB&1h&3600&-&0,06 \%&188 \%& -&7&8&10\\
\hline
LB&3h&&6,51 \%&1,14 \%&7,46 \%&10,38 \%&0&2&6\\
UB&3h&&3654,06 \%&0,00 \%&3,93 \%&701 \%&10&12&14\\
\hline
LB&\cite{dupin2016dual}&&2,86 \%&1,14 \%&2,50 \%&4,27 \%&0&0&7\\
LB&Best&&2,77 \%&1,07 \%&2,50 \%&4,27 \%&0&1&7\\
\hline
\end{tabular}

\end{table}

\subsection{{MILP solving characteristics}}

The characteristics of MILP solving is analyzed with Tables \ref{CplexConv}, \ref {CplexROADEFandExt2}
and \ref{tempsCalcul}.

Tables \ref{CplexConv} and \ref {CplexROADEFandExt2} compare
 the gaps to the BKS for different LB and UB at different stage of the B\&B solving:
LB of the LP relaxation, LB and UB at the root note after cutting plane passes and before branching,
and B\&B computation in time limits of 1h and 3h.
Table \ref{CplexConv} highlights that
the B\&B algorithm is very efficient on the 18 instances where the optimal value is known, 
with high quality LB and UB of the B\&B solving at the root node.

On the most difficult instances,  the UB and the MILP primal heuristics may be of a poor quality,
as shown by Table \ref {CplexROADEFandExt2}.
Actually, the LP relaxations are not useful to  compute primal solutions for instances like B8 and B9.
In these cases,  several production cycles can overlap and the LP relaxation  gives  very few integer variables, specially for last cycles.
Hence, this induces difficulties for MILP primal heuristics, relying on the LP relaxation such as Feasibility Pump, RINS or Local Branching,
and tailored variable fixing heuristics.
This explains two bottlenecks for straightforward MILP solving in the instances with large time windows shown in Table 8:
the LP relation is significantly of a  worse quality and the search of primal solutions is inefficient, compared to the easier instances with  smaller time windows. On one hand, Such difficulties have to be overcome with the matheuristics.
On the other hand, matheuristics can rely on the efficiency of B\&B search for restricted sizes of instances.

We note that the LB obtained with the straightforward B\&B algorithm and with the dual heuristics of \cite{dupin2016dual}, are of a good quality, these dual bounds are reported and denoted "Best LB" in Table 9 for the instances where the optimal value is not proven by the straightforward B\&B computations. These dual bounds improve significantly the ones from the straightforward MILP solving, 
which allows  relevant comparisons with the primal heuristics.

 \begin{table}
\centering
\caption{Comparison to the gaps to the BKS for different LB and UB at different stage of the B\&B solving:
LB of the LP relaxation, LB and UB at the root note after cutting plane passes and before branching,
 and after 1h and 3h computations.
}
\label{CplexROADEFandExt2}

\vskip 0.3cm

\begin{tabular}{|l|cc|ccc|cc|cc|}
\hline
&LP&LP&nod1&nod1&nod1&1h&1h&3h&3h\\
&t &LB&t&LB&UB&LB&UB&LB&UB\\
\hline
B6&4,4&2,59 \%&25,6&0,92 \%&0,45 \%&0,20 \%&0,00 \%&0,06 \%&0,00 \%\\
B7&17,8&4,74 \%&220,4&3,37 \%&-&2,49 \%&0,03 \%&2,02 \%&0,02 \%\\
B8&147&12,23 \%&1,1k&11,82 \%&-&10,81 \%&-&10,81 \%&-\\
B9&167&11,62 \%&1,5k&10,92 \%&- &10,39 \%&- &10,39 \%&-\\
B10&8,3&3,97 \%&36&1,76 \%&0,12 \%&0,44 \%&0,00 \%&0,23 \%&0,00 \%\\
\hline
X11&3,5&1,82 \%&30&0,73 \%&0,85 \%&0,17 \%&0,01 \%&0,10 \%&0,00 \%\\
X12&7,7&3,25 \%&45&1,20 \%&0,39 \%&0,37 \%&0,01 \%&0,12 \%&0,00 \%\\
X13&5,5&3,11 \%&131&1,96 \%&- &1,38 \%&0,16 \%&1,14 \%&0,01 \%\\
X14&6,3&3,56 \%&64&2,08 \%&1,75 \%&1,34 \%&0,07 \%&1,12 \%&0,05 \%\\
X15&6,3&3,95 \%&25&0,86 \%&0,15 \%&0,09 \%&0,00 \%&0,05 \%&0,00 \%\\
\hline
\end{tabular} 
\vskip 0.3cm

\begin{tabular}{|l|cc|ccc|cc|cc|}
\hline
&LP&LP&nod1&nod1&nod1&1h&1h&3h&3h\\
&t &LB&t&LB&UB&LB&UB&LB&UB\\
\hline
B6Ext2&121&9,34 \%&663&8,4\%&-&7,97 \%&-&7,59 \%&5,23 \%\\
B7Ext2&251&10,63 \%&738&10,0\%&-&9,70 \%&-&9,70 \%&10,62 \%\\
B8Ext2&174&14,87 \%&1473&14,5\%&-&13,62 \%&-&13,62 \%&-\\
B9Ext2&390&13,30 \%&3,6k &12,6\%&-&12,60 \%&-&12,15 \%&-\\
B10Ext2&78&13,26 \%&801&12,3\%&-&11,91 \%&-&11,91 \%&- \\
\hline
X11Ext2&153&9,44 \%&2,1k&8,5\%&-&8,20 \%&- &8,20 \%&-\\
X12Ext2&109&11,09 \%&974&10,0\%&-&9,83 \%&-&9,62 \%&9,33 \%\\
X13Ext2&195&12,41 \%&2,28k&11,0\%&-&10,37 \%&-&10,37 \%&-\\
X14Ext2&152&11,88 \%&1,97k&10,6\%&-&10,14 \%&-&10,14 \%&-\\
X15Ext2&197&13,40 \%&896&11,8\%&-&11,29 \%&- &11,29 \%&13,78 \%\\
\hline
\end{tabular}
\end{table}

Table \ref{tempsCalcul} compares the termination time of B\&B to optimality on the smallest instances:
     without and with the exact pre-processing of \cite{dupin2018dual},
     the influence of  warmstarting the B\&B algorithm  with the optimal (or best known) solution.
 The exact pre-processing of \cite{dupin2018dual}, tightening time windows thanks to constraints CT5 and CT8-11,
has an strong influence in the computation times to optimality.
 Warmstarting has less influence for the small instances.
Warmstarting is relevant
 when the search of primal solutions is inefficient, which is the bottleneck of MILP solving noticed for the most difficult instances.
For the MILP computations inside the MIP-VND matheuristic, warmstarting is also relevant, especially 
for the largest neighborhoods.

\subsection{Optimizing around an initial solution}

In the instances from the ROADEF Challenge,  no initial solution was given.
To analyze the impact of having an initial solution and minimizing with stability costs,
we considered as initial solutions the ones given by the matheuristic of Section 5.1 without the repairing post-processing.
Such initial solutions are not always feasible, and are of a medium quality.

Adding penalization to a baseline solution in the objective function changes  the MILP convergence characteristics. 
Guiding the solution search around the baseline solution improves very significantly
the solving time, as shown in Table \ref{tempsCalcul}.
The higher are the stability coefficients $\mathbf{C}^{pen}_{i,k,w}$,
the more the B\&B algorithm is accelerated.
Many solutions with similar ROADEF costs exist  which was already noticed for the Challenge, especially in \cite{gardi}.
Similarly with symmetries in the  B\&B tree search, it induces that few cut-off are processed, and the optimal
computations explore many nodes where  primal solutions with similar costs are found.
On the contrary, the tree search is guided around the LP relaxation with high stability costs,
with fewer nodes in the  B\&B tree to explore.
As is, it can be used to repair efficiently infeasible solutions.

\begin{figure}[ht]
   \begin{minipage}[c]{.48\linewidth}
\centering    
   
      \includegraphics[angle=0, width=.99\linewidth]{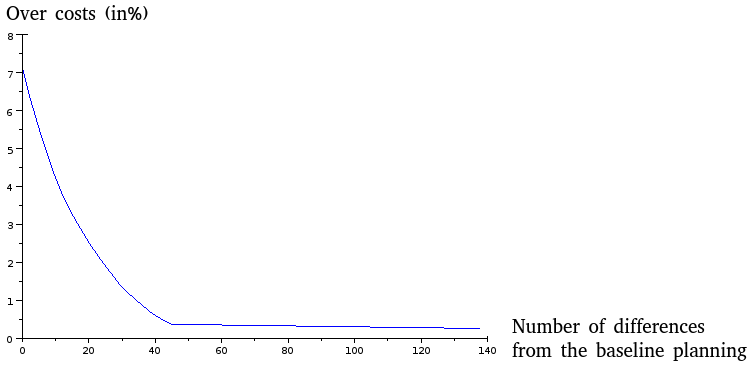}
{Instance B7-3-120}
   \end{minipage} \hfill
   \begin{minipage}[c]{.48\linewidth}
\centering       
      \includegraphics[angle=0, width=.99\linewidth]{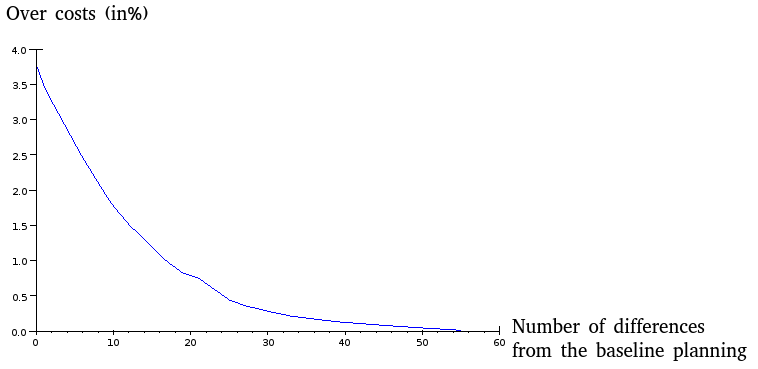}
{Instance X11-3-120}
   \end{minipage}
      \caption{Pareto frontiers 
for the best compromises cost/stability: the stability indicator is measured 
as the number of differences from a baseline solution (computed with the heuristic of section 5.3). 
}\label{initPareto}
 \end{figure}

These  characteristics allow  to  consider an $\varepsilon$-constraint method 
 to draw the Pareto curve of non dominated solutions in the bi-objective optimization cost/stability
 (\cite{laumanns2006efficient}).
The granularity for the $\varepsilon$-constraint method can be set to $\varepsilon =1$ without
granularity errors,
 using the integer objective functions for stability penalization.
Figure \ref{initPareto} presents such Pareto frontiers for instances \verb!B7-3-120! and \verb!X11-3-120!.
The quality of the baseline solution has a consequent impact for the Pareto frontiers,
with a poor quality solution like the ones given by the constructive heuristic of section 5.1,
an important trade-off exist between cost and stability.
The price of stability is higher than the price of robustness shown in \cite{dupin2016meta}.

\begin{table}

\centering
\caption{Comparison to the gaps to the BKS for the solutions of matheuristics, to Cplex solving and the dual bounds from \cite{dupin2016dual}.
}\label{tabPANheur}
\vskip 0.3cm
Results for the instances where the optimal value is known. (over 18 instances )
\vskip 0.2cm

\begin{tabular}{|l|c|cc|cc|ccc|}
\hline
&$\bar{t}$&$\bar{g}$&$\sigma{g}$&Q1&Q3&$N_{0.01}$&$N_{0.05}$& $N_{1}$\\
\hline
Cplex 1h&217&0,00 \%&0,00 \%&0,00 \%&0,00 \%&18&18&18\\
\hline
RRF&238&0,07 \%&0,23 \%&0,01 \% \%&0,01 \%&11&16&18\\
RRF+1VND&270&0,01 \%&0,05 \%&0,00 \%&0,00 \%&17&17&18\\
RRF+VND&313&0,01 \%&0,05 \%&0,00 \%&0,00 \%&17&17&18\\
\hline
Simpl&59&0,08 \%&0,24 \%&0,00 \%&0,02 \%&10&15&18\\
Simpl+1VND&93&0,01 \%&0,05 \%&0,00  \%&0,00 \%&17&17&18\\
Simpl+VND&137&0,01 \%&0,05 \%&0,00  \%&0,00 \%&17&17&18\\
\hline
CMSA&63&0,03 \%&0,07 \%&0,01  \%&0,03 \%&6&16&18\\
CMSA+1VND&96&0,01 \%&0,05 \%&0,00  \%&0,00 \%&16&17&18\\
CMSA+VND&133&0,01 \%&0,05 \%&0,00  \%&0,00 \%&16&17&18\\
\hline
Best VND&&0,01 \%&0,05 \%&0,00 \%&0,00 \%&17&17&18\\
\hline
\end{tabular}

\vskip 0.3cm
Results for the 22 most difficult instances (B8, B9, and Ext2 and Ext3 datasets).
\vskip 0.2cm


\begin{tabular}{|l|c|c|cc|ccc|c|}
\hline
&$\bar{t}$&$\bar{g}$&Q1&Q3&$N_{0.01}$&$N_{0.05}$& $N_{1}$&$N_F$ \\
\hline
Best LB&12k&9,20 \%&7,39 \%&11,17 \%&0&0&0&0\\
\hline
UB Cplex 1h&3,6k&-&3,71 \%&-&0&0&0&17\\
UB Cplex 3h&10,8k&-&2,63 \%&1842 \%&2&2&3&10\\
\hline
RRF&6,6k&2,10 \%&0,07 \%&3,55 \%&0&5&7&0\\
RRF+1VND&7,9k&0,64 \%&0,06 \%&0,39 \%&2&5&17&0\\
RRF+VND&15k&0,05 \%&0,00 \%&0,04 \%&11&17&22&0\\
\hline
Simpl&0,9k&28,15 \%&11,51 \%&40,45 \%&0&0&22&0\\
Simpl+1VND&2,2k&2,00 \%&0,06 \%&2,35 \%&1&5&8&0\\
Simpl+VND&10,6k&0,11 \%&0,00 \%&0,06 \%&9&14&1&0\\
\hline
CMSA&0,9k&28,16 \%&12,30 \%&37,88 \%&0&0&22&0\\
CMSA+1VND&2,2k&2,01 \%&0,08 \%&2,71 \%&1&3&8&0\\
CMSA+VND&10,1k&0,07 \%&0,01 \%&0,07 \%&6&14&0&0\\
\hline
Best VND&&0,00\% \%&0,00 \%&0,00 \%&21&21&22&0\\
\hline
\end{tabular}

\vskip 0.3cm
Results over all the 50 instances.
\vskip 0.2cm

\begin{tabular}{|l|c|c|cc|ccc|c|}
\hline
&$\bar{t}$&$\bar{g}$&Q1&Q3&$N_{0.01}$&$N_{0.05}$& $N_{1}$&$N_F$ \\
\hline
LB Best&-&1,77 \%&0,01 \%&3,20 \%&18&19&25&0\\
\hline
Cplex 1h&&-&0,00 \%&-&25&26&28&17\\
Cplex 3h&&-&0,00 \%&8,98 \%&28&30&32&10\\
\hline
RRF&6,6k&0,97 \%&0,01 \%&0,32 \%&17&28&43&0\\
RRF+1VND&8k&0,29 \%&0,00 \%&0,14 \%&27&30&45&0\\
RRF+VND&19,7k&0,03  \%&0,00 \%&0,02 \%&36&42&50&0\\
\hline
Simpl&0,9k&12,50 \%&0,02 \%&17,21 \%&10&15&26&0\\
Simpl+1VND&2,4k&0,89 \%&0,00 \%&0,20 \%&27&31&42&0\\
Simpl+VND&6,8k&0,05 \%&0,00 \%&0,03 \%&35&40&49&0\\
\hline
CMSA&0,9k&12,47 \%&0,03 \%&22,13 \%&6&16&28&0\\
CMSA+1VND&2,4k&0,89 \%&0,00 \%&0,28 \%&26&29&42&0\\
CMSA+VND&9,7k&0,04 \%&0,00 \%&0,03 \%&31&40&50&0\\
\hline
Best VND&&0,01 \%&0,00 \%&0,00 \%&46&46&50&0\\
\hline
\end{tabular}

\end{table}

\subsection{{Constructive matheuristics}}

With the results of Table \ref {CplexROADEFandExt2}, the first challenging issue is to
compute accurately primal solutions for the most difficult instances.
The numerical  results of constructive heuristics are highlighted in Table \ref{tabPANheur}.
Simplified MILP computations of section 5.1, the CMSA and the  R-R\&F heuristics succeeded to provide feasible solutions for all types of instances, without failures, contrary to straightforward Cplex solving in one or three hours.
This was the first motivation to design these primal heuristics.

The MILP formulation of section 5.1 is easier for the MILP primal heuristics,
it accelerates significantly the search of primal solutions.
For the difficult  instances like B8 and B9, feasible solutions were found in 15 minutes including the repairing post-processing,
as shown in Table \ref{tabPANheur}.
The CMSA algorithm requires a similar computation time than the simplified MILP, with a similar (slightly better) quality of solutions.
 R-R\&F approach gives significantly better solutions, but requiring longer computation times to converge to a feasible solution.
 
 We note that the computation times from 15 minutes to two hours for the constructive heuristics were not optimized, the objective was 
 to solve accurately the instances without over-fitting.
 Contrary to the  R-R\&F algorithm where the time limit in MIP solving is a bottleneck, 
 the computation times can be decreased for the  CMSA and the simplified MILP approaches.
Such computation times are reasonable for the industrial application as a first phase before local search improvements.
At this stage, the goal was to be able to compute accurately feasible solutions for all the instances, it is reached. The VND matheuristic can be initialized with feasible solutions
from the three types of constructive matheuristics.

\subsection{POPMUSIC-VND matheuristics} 

Table \ref{vndLocalMin} compares the quality of the local minimum considering each type of single neighborhoods separately in the MILP-VND local search.

Neighborhoods $\N_{(0,1)}^{TW}$ include the  neighborhood to move one single outage from one week, considered by many approaches for the ROADEF Challenge.
$\N_{(0,1)}^{TW}$ includes also 1-translation neighborhoods, translating from one week the maintenance planning.
Table \ref{vndLocalMin}  shows that it induces a poor quality of local minimums.
Many local minimums exist with neighborhoods $\N_{(0,1)}^{TW}$  and also for the smaller neighborhoods.
The quality of local minimums is improved increasing the radius with $\N_{(0,k)}^{TW}$ neighborhoods.
However, significant over-costs remain.
It justifies to consider other types of neighborhoods in a local search method for the problem.
We note that ``funnel" neighborhoods $\N_{(a,b)}^{TW}$ with $a>0$ are interesting instead of using only $\N_{(0,b)}^{TW}$ neighborhoods.

Cycle neighborhoods $\N_{k,k'}^{cycles}$ allow to swap 
the order of outages, which is unfeasible with small $\N_{(a,b)}^{TW}$
neighborhoods with spacing constraints CT14-CT15. Significant improvements are observed in Table \ref{vndLocalMin}, but it requires more computation time than the other neighborhoods.
Swap moves are also possible with $\N_{I}^{units}$ neighborhoods, when subset $I$ contains units in a site defined by constraints CT14-CT15.
Local minimums are of an excellent quality considering unit re-optimization with $|\II|=1$, with MILP sub-problems that converges very quickly (order of one second).

Table \ref{tabPANheur} shows the final results considering in the POPMUSIC-VND neighborhoods $\N_{(1,3)}^{TW}$,  $\N_{k,k}^{cycles}$
and $\N_{I}^{units}$ with $|\II|=5$.
The VND local search starting with the CMSA solutions is very efficient, improving quickly primal solutions and
requiring few VND iterations to compute a local minimum for all neighborhoods.
The resulting matheuristic is more efficient in the ratio improvement of solution/computation times than the R-R\&F heuristic.
Furthermore, comparing the local extrema given by the VND to the optimal solutions of small instances,
we had only one  example where the  VND induces over costs more than the tolerance gap of Cplex parametrized to $0,01 \%$.
The tolerance gap induces sometimes that the VND solutions were better than the ``optimal'' solutions with the terminantion of the B\&B algorithm, within the $0,01 \%$ tolerance gap.

 This illustrates that VND-MILP allows to define easily neighborhoods induced by operational expertise.
 The genericity of MILP modeling and resolution  avoid specific implementations. 
 An application of MILP-VND  is to design and test neighborhoods with few implementation efforts, to select neighborhoods
 that should be implemented  properly  in a local search as in \cite{beImproved,gardi}.
 These results give some insights to explain some results of local search heuristics used for the Challenge.

\subsection{Explanations of some results of the Challenge}

This section give some insights provided by our intermediate results to analyze the design of previous approaches, in mathematical modelling, and also for the operators  of local search heuristics.

\subsubsection{MILP formulations and efficiency of B\&B solving}

Our formulation is  similar to deterministic version of the MILP formulation in  \cite{lusby}. The major  difference  is in the definition of binary variables  $d_{i,k,w}$: binaries $x_{i,k,w}$ defined in \cite{lusby} are equal to $1$ if and only if outage beginning week for cycle $(i,k)$ is exactly $w$.
Actually, these definitions can be seen as isomorphic (as in the analysis of \cite{ucp}) with $x_{i,k,w} = d_{i,k,w}- d_{i,k,w-1}$ and $d_{i,k,w} = \sum _{w'\leqslant w} x_{i,k,w'}$. We note that this transformation applies also for the matheuristic approach \cite{jost}.
If the MILP formulations are polyhedrally equivalent, it can have an influence in the efficiency of MILP solvers, and different activation of cuts with Cplex, as in\cite{ucp}. 
This is not crucial in the master problem of the Bender's decomposition  in \cite{lusby} or in the MILP model of scheduling constraint in \cite{jost} as this problems are quickly solved by Cplex.
This had more impact for the exploration of our neighborhoods, we observed than our formulation provides better branching (well-balanced branching similarly with SOS1 branching), whereas the formulation of \cite{lusby} was slightly better for the efficiency of the MILP primal heuristics.

Adding continuous variables with the production level makes the MILP solving with Cplex difficult, as shown in Table 8 and in the Appendix A. 
This justifies the Bender's decomposition approach to tackle the problem with many stochastic scenarios, as in \cite{lusby}, this would be also a perspective to generate only the CT6 constraints that are useful.
The difficulty met with an approximate description of CT6 constraints using only continuous variables
makes the exact MILP formulation of \cite{Jonc10} intractable even for small sizes of instances.
Lastly, we mention that solving the LP with fixed maintenance dates, which corresponds to $\N_{(0,0)}^{TW}$ computations, are solved easily
by Cplex (in less than one second). There is no need anymore for decomposition approaches like price decomposition,
as used in ancient work  with less efficient LP solvers and hardware like in \cite{Dub05}.

\subsubsection{Extended formulations for CT6 constraints?}

Several approaches used extended formulation and column generation, as in  \cite{Roz12,griset2018methodes},
 to deal with an exact description of the CT6 "stretch" constraints.
The Appendix A give results that show that CT6 constraints have few impact in the quality of the final solutions: 
a little over-cost is given repairing the stretch constraints from an optimal solution relaxing the CT6 constraints,
and very local modifications allow to decrease significantly the over-costs to negligible over-costs.
Actually, this can be analyzed and explained numerically: stretch constraints are not often active, it is more costly to have a non maximal T2 production as it is cheaper. These situations can be unavoidable for feasibility and time windows constraints. In such cases, it concerns few weeks in the planning, and the local  optimization is efficient to incorporate the CT6 constraints and improve the cost.
We note that the column generation approaches are more limited than our matheuristics:
\cite{Roz12} was not able to implement the Branch\&Price resolution because of time limit constraints, with 40 minutes to solve the LP relaxation of the extended model, and  \cite{griset2018methodes} restricted the time horizon to 3 years instead of 5 in the Challenge.
Willing to face a 5 year time horizon, with the large time windows like datasets Ext2 and Ext3, the capacity of Branch\&Price resolution would be still limited nowadays.

\subsubsection{2-stage decomposition approaches}
 
The 2-stage decomposition is a natural approach, taking separately  
  maintenance planning  and  production decisions having fixed the other decisions.
  For the production decisions, sub-problems are independent for all the stochastic scenarios like in the Bender's decomposition approach \cite{lusby}.
  Many heuristic approaches used this 2-stage decomposition: \cite{Ang12,brandt,Bra13,Gav13,Godskesen,jost}.
  Having the maintenance and refueling decisions fixed, it makes  easier  to compute the productions.
For a maintenance planning, it computes the production, indicates if the planning is feasible and what is the cost, using a LP or a greedy algorithm.
 In our simplified model, it corresponds to  $\N_{(0,0)}^{TW}$ computations, which are very fast.
Modifying the maintenance decisions were in these approaches dealt by very local transformations: translating the maintenance planning of one weeks for consecutive outages of a T2 power plant, or modifying one outage by one week. 
Such neighborhoods are smaller than our $\N_{(0,1)}^{TW}$ neighborhoods, with many local optimums and a poor quality of local minima as shown in Table 13.
This is an explanation of the over-costs given by approaches \cite{Ang12,brandt,Bra13,Gav13,Godskesen,jost}.

Another difficulty of such approaches are that the CP or MILP computations are more time consuming than straightforward local search approaches,
it induces to explore less solutions in the solving time.
In our VND, the MILP explorations cover larger neighborhoods than just computing the costs.
These MILP iterations are more costly than $\N_{(0,0)}^{TW}$ computations.
The VND has the advantage of furnishing directions of optimizations, which helps the coordination phase to modify the maintenance planning.
On the contrary, 2-stage decompositions modify the maintenance planning problem blindly with random moves from the current planning. 
This is another explanation of the limitations of such approaches, while the optimal exploration with $\N_{(0,1)}^{TW}$ is more efficient to accomplish a similar task.
We note that motivations of the work  \cite{barty2014sensitivity} was also to give directions of re-optimization for a 2-stage decomposition approach.

\subsubsection{Local search without decomposition} 

 The analyses of  neighborhoods with VND shows that the ``Winter Jump" neighborhoods are crucial to avoid being struggled in local minimum, as shown in Table 13. The neighborhoods "k-MoveOutagesConsecutive" in \cite{gardi}, as mentioned in  section  \ref{LSneighborhood}, incorporate these winter jumps.  
We note that the "k-MoveOutagesRandom" and "k-MoveOutagesConstrained"  in \cite{gardi} contains similar neighborhoods than our $\N_{(a,b)}^{TW}$ and $\N_{k,k'}^{cycles}$ neighborhoods. The benefit of combining these three types of neighborhoods may have a similar influence than our Table 13 to have
local minimums of an excellent quality. Such choice of neighborhoods is a key component to design a local search heuristic for this problem.
This result was analyzed with matheuristics, the implementation of  \cite{gardi} allows to have quicker moves and thus a more scalable approach.
The approach of \cite{gardi} is able to explore aggressively a lot of solutions, an is efficient also within 15 minutes of resolution time,
contrary to our matheuritics which require heavier computations.

We note that  \cite{beImproved} did not publish elements on the final design of neighborhoods.
They mentioned having used only neighborhoods which are smaller than $\N_{(1,0)}^{TW}$ in the first phase, like the 2-stage decomposition approaches.
The improvements in the quality results of  \cite{beImproved} between the qualification phase and the final result
are due to the choice of other neighborhoods. 
The over-costs was in average of $2,8\%$ for the qualification phase, whereas they had  6 best solutions and very few over-cost in general for the final results. 
This is another proof of the crucial impact of the design of neighborhoods for this problem.
It explained also the worst quality of solutions obtained by 2-stage decomposition  approaches.

\subsection{Contributions for the industrial application}

Our results give also contributions for the industrial application.
A first result is the validation of the design of the heuristic of \cite{gardi}, which explains its quality.
This was not obvious after the Challenge why such approaches were the most efficient, and practitioners may mistrust optimization approaches without
guarantee of optimality or dual bounds. We note that dual bounds are given now in \cite{dupin2018dual}, which validates the quality of the solutions furnished for the Challenge. With this work, one may trust the design of such heuristics to give good solutions for instances 
that are met in the operational application.
We note that experts could have the intuition that ''winter jumps" may be en efficient neighborhood, the results of \cite{gardi} and Table 13
justify it is a crucial point.
The following of this section highlights the impact of intermediate results for the industrial application.

\subsubsection{Relevancy of simplification hypotheses}

Before the Challenge, the operational resolution of the problem in EDF contains the approximations of our model mentioned in section \ref{simplSec}, we refer to \cite{Kh07,griset2018methodes}.
This paper contains even  more constraints and less approximations, relaxation of constraints CT6,CT12 but also CT14-CT21
and also no fuel optimization were common assumptions.
Another heuristic reduction was to tighten the constraints of  time windows, to ease the computations (or make it possible in reasonable time). Looking to the instances of the Challenge in Table 3, the heuristic reduction apply to all the instances except B8-B9 (and thus the reproduced instances of this study).
One  reason motivating the Challenge, was to give the problems without the approximations and simplifications used in  their homemade resolution, and to analyze if the approximations are relevant.

The aggregation of production time steps and  stochastic scenarios were also used in \cite{Roz12}.
It induced  an overall over-cost of $0,5 \%$ compared to the best solutions, mainly the one from  \cite{gardi} without such aggregation.
This gap main include also over-costs due to the non termination of the Branch\&Price algorithm, with hierarchic computations in \cite{Roz12}.
In Table 6, the impact of the heuristic tightening of time windows in a pre-processing phase, can be evaluated to $1\%$, which is huge.
Our matheuristics and the local search of \cite{gardi} are still efficient with large time windows, the reduction is not required anymore for 
solving capabilities, and this hypothesis seems to have the highest impact in terms of over-costs.
Furthermore, the impact of relaxing CT6 constraints is negligible, as analyzed in Appendix A and mentioned in section 7.7. 
This work gave the impact of simplification hypotheses, the assumptions taken in this paper are a very good compromise in minimizing the over-costs and minimizing the computational difficulties, which is useful for the operational application.

\subsubsection{Towards dynamic optimization}
 
 The impact of the optimization with large time windows, as previously highlighted, 
 has also incidences in the dynamic re-optimization process considering stability costs.
 The Figure  \ref{initPareto} illustrated that the initial baseline solution
 can have a huge impact in the over-costs while searching for a solution close to the baseline solution.
 In these cases, the over-costs may be of $5\%$, whereas starting from a very good solution  has less impact.
 By comparison, the over-costs of having more robust solution was in order of $0,1-0,5\%$ in \cite{dupin2016meta}.
 For this applications, it is interesting to have a maximal flexibility to optimize the latest outages with $k \geqslant 3$,
 so that the stability considerations would start with a good solution for these outages when the considered outage  have to be fixed for stability while term is approaching.

\section{Conclusions and perspectives}

This paper investigated matheuristic approaches to optimize refueling and maintenance scheduling  of nuclear power plants. 
Instances with time window constraints only for the first maintenance outages
induce difficulty for state-of-the art MILP solvers.
Several constructive matheuristics use decompositions  to have more accurately feasible solutions solving with  smaller problems in heuristic decomposition schemes.
Once feasible solutions are built, a POPMUSIC-VND matheuristic allows to have accurately solutions of an excellent quality,
 with standard computers in reasonable time limits.
Using matheuristics allows also a genericity in the objective functions, to derive a bi-objective optimization to find the best compromises solutions between the planning stability and the financial costs.

Furthermore,  intermediate results help to understand better the problem characteristics and
some ranking after the EURO/ROADEF 2010 Challenge.
This article justified some approximations made by \cite{lusby,Roz12,jost} to have tractable MILP formulations to face the large size of real size instances:
aggregation of production time steps to week, at least partial aggregation of stochastic scenarios.
The relaxation of the constraints CT6 are also justified thanks to the matheuristics.
The matheuristic methodology, from MILP formulation work to the design of constructive and local search approaches, allows to analyze the impact of neighborhoods in the quality of local minima.
This analysis helps to understand why many decomposition approaches failed for the challenge, because of too
short neighborhoods. Our matheuristic study explains why approaches developed in 
\cite{gardi,beImproved}
for the Challenge were more efficient: using no time consuming  LP, MILP or CP  computations, more time could be devoted 
to explore solutions, and the use of variable and larger neighborhoods than $\N_{(0,1)}^{TW}$ for a better exploration.
Especially, the "Winter Jump" neighborhoods are crucial.

These results offer several perspectives for the industrial application.
If the design of approaches \cite{gardi,beImproved} are validated by this study,
perspectives to improve other approaches are discussed in details in Section 7.7.
The impact of the reduced time windows is shown in this study to have an enormous impact,compared to the other simplification hypothesis.
We note that the heuristic reduction is an hypothesis to ease computations, the solving capabilities of our matheuristics and the frontal local searches  \cite{gardi,beImproved} do not need such simplification to run in reasonable time.
Furthermore, we note than fixing local optimization around the known solution in the dynamic optimization application with the monthly re-optimization has similar effects, our bi-objective optimization work 
recommends to care about it in the real-life application.
Another perspective is to consider other objective functions: 
maximizing robustness to lessen financial risks,
or sustainable development issues like minimizing nuclear wastes or minimizing carbon emissions in regards to financial over-costs. This works illustrates that many solutions with similar costs exist, which was known already after  \cite{gardi}. One may  choose a maintenance planning among the best 
ones with the cost optimization given for the Challenge
and looking for a better trade-off using several objective functions.

 \bibliographystyle{plain}    
\footnotesize
 \bibliography{biblioNuc}
\normalsize

\newpage

\section*{{Appendix A: relaxation of constraints}}

In this appendix, we justify the  relaxation  of modulation constraints CT12,
and decreasing profile constraints CT6.
Such constraints were often relaxed in the approaches for the  EURO/ROADEF 2010 Challenge,
inducing more complexity and difficulty in the optimization modeling and solving.
This appendix provides justifications, regarding the industrial application
and some numerical analyses.

\subsection*{{A.1 Modulation constraints CT12}}

In this paper, modulation constraints, denoted CT6 in the Challenge description, are relaxed.
In the Challenge, it is formulated as a maximal  volume of energy generated 
when a unit $i$ is not producing at the maximal power $\mathbf{\overline{P}}_{i,w}$ at time step $w$,
the cumulated energy being counted for all cycle $k$ when the fuel level is superior to $\mathbf{Bo}_{i,k}$.
When the fuel level is inferior to $\mathbf{Bo}_{i,k}$, the production is at the maximal power defined by the CT6 decreasing profile.

\begin{figure}[ht]
      \centering
      \includegraphics[angle=0, width=.99\linewidth]{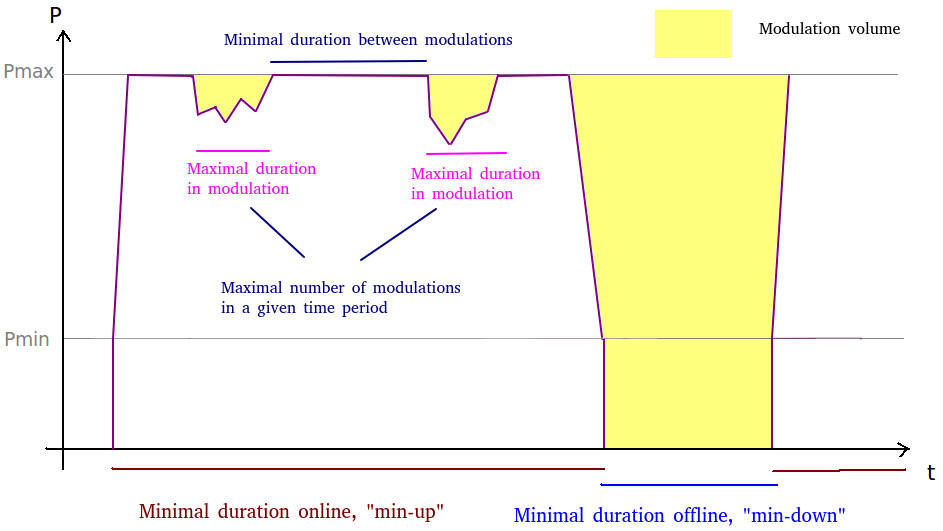}

      \caption{Illustration of modulation constraints for nuclear power plants}\label{defModConstr}
   
\end{figure}

Actually, such modulation constraints are approximated and aggregated  from technical constraints of nuclear power plants.
The fine grain  technical constraints are as following, similarly to \cite{Tak05,ucp}, and illustrated Figure \ref{defModConstr} :

\begin{itemize}
 \item \textit{Min-up/min-down} constraints:  every unit $u$ has a minimum up time $\Delta^{on}_{u}$ online and a minimum down 
time $\Delta^{off}_{u}$ offline.
  \item \textit{Max duration with a non maximal power} constraints: if a nuclear power plant is online, there is a maximal duration
  to have modulated power, i.e. non maximal power. 
  
  \item \textit{Min duration between consecutive modulations}: between consecutive modulations, there is a minimal duration
  to have  maximal power. 
  
  \item \textit{Max number of modulation} constraints: for each day, there is a maximal number of
  periods where the production is not maximal.
    \end{itemize}

If a T2 unit is online, the three last constraints induce a maximal volume of modulated power.
Actually, the CT12 modulation constraints consider that the T2 units are always on-line in a production cycle.
This last assumption has  its justification as T2 production is less expensive than the T1 production.
However, modulation may be necessary with  low demands (lower than the available T2 production).
In such cases, T2 units can be set offline with the min-up/min-down constraints,
defining a volume of modulation that is not bounded.
In these cases, there will be minimal durations offline with {Min-up/min-down} constraints.
These extreme modulations happen in Figure \ref{prodStretch} using the model of section 4.

\subsection*{{A.2 Decreasing profile for T2 units CT6}}

\begin{figure}[ht]
      \centering
      \includegraphics[angle=0, width=.97\linewidth]{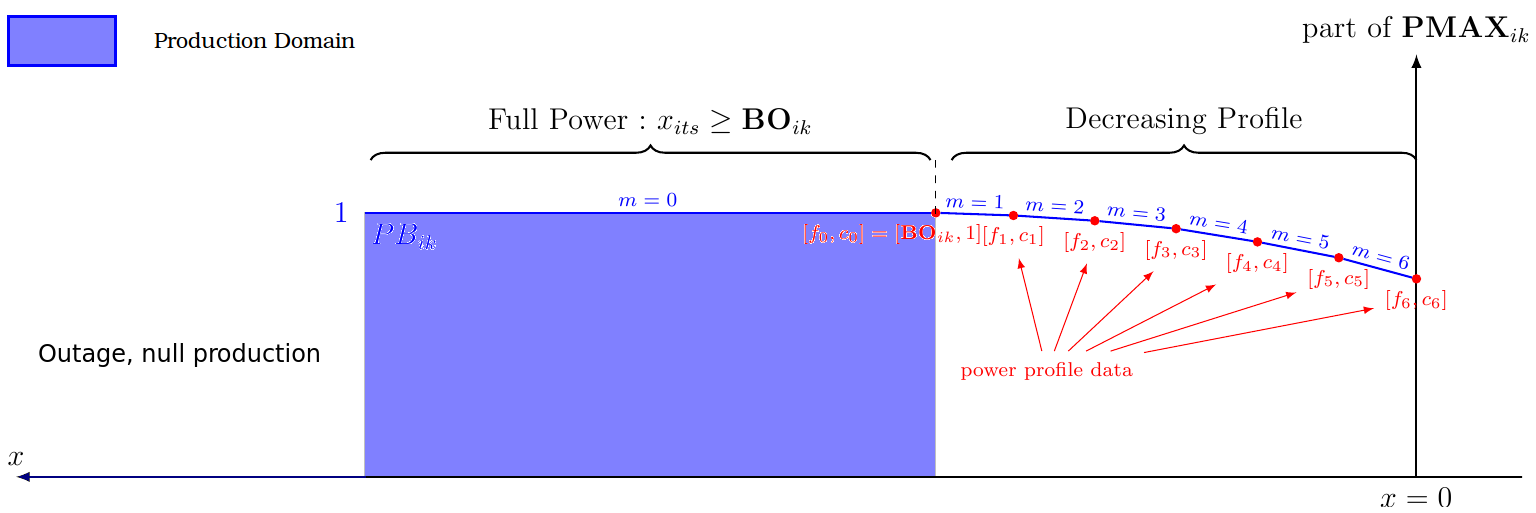}
      \caption{Illustration of the production domain for T2 power plants in the Challenge ROADEF}\label{prodRoadef}
      \vskip 0.1cm
      \includegraphics[angle=0, width=.97\linewidth]{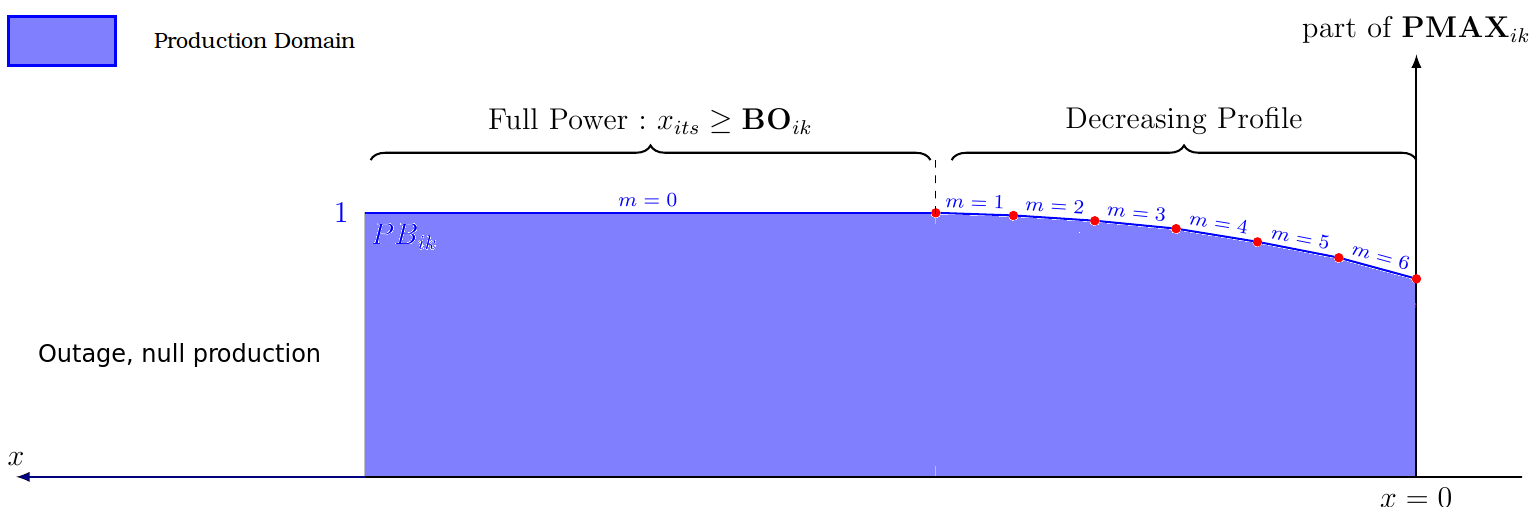}
      \caption{Illustration of the production domain  for T2 power plants with light CT6 constraints}\label{evo1}  
\end{figure}

In the Challenge, the  ``stretch decreasing profile" CT6 are defined when the fuel stock level
of a T2 unit $i \in \II$ is inferior to the level $\mathbf{Bo}_{i,k}$ in the cycle $k \in K_i$.
The production of $i$ is deterministically defined, following a decreasing profile,  piece-wise linear function of the stock level, as  in Figure \ref{prodRoadef}. More precisely:

\vskip 0.3cm

\textbf{CT6, "stretch" constraints}: 
  During  every time step $w\in \WW $ of the production campaign of cycle $k \in K_i$,
if the current fuel stock of plant $i \in \II$ is inferior to the level $\mathbf{Bo}_{i,k}$ , the production of $i$ is deterministic,
following a decreasing profile,  piece-wise linear function of the stock level, as  in Figure \ref{prodRoadef}.  
$m \in \MM_{i,k}= [\![1,\mathbf{Np}_{i,k}]\!]$ define the points of the  CT6 profile ending the cycle $(i,k)$.
The points are denoted $(\mathbf{f}_{i,k,m},\mathbf{c}_{i,k,m})$ for each cycle  $(i,k)$,
$\mathbf{f}_{i,k,m}$ denote the fuel levels,
$\mathbf{c}_{i,k,m} \in [0,1]$ is defined as the loss coefficient in power compared to the maximal power.

\vskip 0.3cm

We  will consider a lighter constraint , imposing only the UB constraints as illustrated in Figure \ref{evo1}.
This MILP formulation was already introduced in  \cite{dupin2018dual}.
New fuel variables $x_{i,w}$ are introduced, representing the residual fuel at week $w$ for T2 unit $i$.
These variables require the following  constraints to
enforce $x_{i,w}$ to be the residual fuel:
\begin{equation}\label{defVarStretch}
 \forall  i,k,w,\; \; \; x_{i,w} \leqslant x_{i,k}^{i} -  \sum_{w'\leqslant w} \mathbf{F}_{w'} \:  p_{i,k,w'}
+ M_i \;  (1 - d_{i,k,w} + d_{i,k-1,w}) 
\end{equation}
where $M_i = \max_k \mathbf{\overline{S}}_{i,k}$, such that $M_i$  verifies $x_{i,t} \leqslant  M_i$. 
Indeed, if $d_{i,k,w} - d_{i,k-1,w}=1$, week $w$ happens in cycle $k$, the active constraint is
 $x_{i,t} \leqslant x_{i,k,s}^{i} -  \sum_{w'\leqslant w} \mathbf{F}_{w'} \:  p_{i,k,w'}$,
otherwise we have  $x_{i,t} \leqslant  M_i$, which is always true.
The UB of the CT6 constraints, as illustrated Figure \ref{evo1}, are given with: 

\begin{equation}\label{ctStretch1}
\forall  i,k,w,m>0, \; \; \;  \frac {p_{i,k,w}} {\mathbf{\overline{P}}_i^t} \leqslant  
\frac{\mathbf{c}_{i,k,m-1} - \mathbf{c}_{i,k,m}}{\mathbf{f}_{i,k,m-1} - \mathbf{f}_{i,k,m}} (x_{i,w} - \mathbf{f}_{i,k,m}) + \mathbf{c}_{i,k,m}
\end{equation}
If  (  $\mathbf{c}_{i,k,m},\mathbf{f}_{i,k,m-1}$) do not depend on indexes  $k$,
we  can have an equivalent MILP formulation using aggregated constraints, 
writing the  constraints for the global production power $\sum_k {p_{i,k,w}}$:
\begin{equation}\label{ctStretch2}
\forall  i,w,m>0 \;\;\; \sum_k \frac {p_{i,k,w}} {\mathbf{\overline{P}}_{i,w}} \leqslant  \frac{\mathbf{c}_{i,m-1} - \mathbf{c}_{i,m}}{\mathbf{f}_{i,m-1} - \mathbf{f}_{i,m}} (x_{i,w} - \mathbf{f}_{i,m}) + \mathbf{c}_{i,m}
\end{equation}


Adding the lighter CT6 constraints, it makes MILP solving difficult.
Computations were not possible for the  datasets B and X, inducing memory errors.
We note that the aggregated formulation (\ref{ctStretch2}) provided improvements in the computation times compared to (\ref{ctStretch1}), but the
formulation is still untractable for the real size instances.
 It can be explained in terms of number of variables, the extension adds $I \times W$ continuous  variables   $x_{i,w}$,
whereas $I \times W \times K$ continuous  variables were already in the model with T2 productions $p_{i,k,w}$.
The main difference appears in the number of constraints, there were mainly $I \times W \times K$ constraints in the MILP of Section 4 with constraints (\ref{PANcoupling}),
 (\ref{defVarStretch}-\ref{ctStretch2}) require to add  $I \times W \times (K+N_p)$ constraints,
 whereas (\ref{ctStretch1}-\ref{defVarStretch}) require to add  $I \times W \times (N_p+1) \times K$ constraints.
This slows down very significantly the LP computations, when  computable.

\begin{figure}[ht]
      \centering
      \includegraphics[angle=0, width=.94\linewidth]{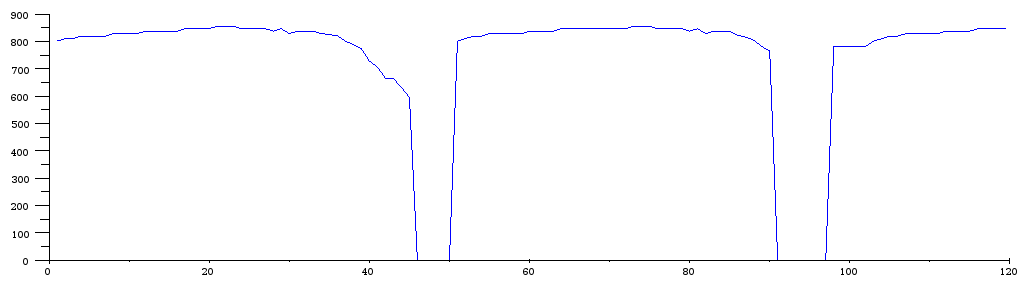}
     \vskip 0.1cm
      
      {Solutions with constraints (\ref{ctStretch1}) }
      \vskip 0.1cm
      \includegraphics[angle=0, width=.94\linewidth]{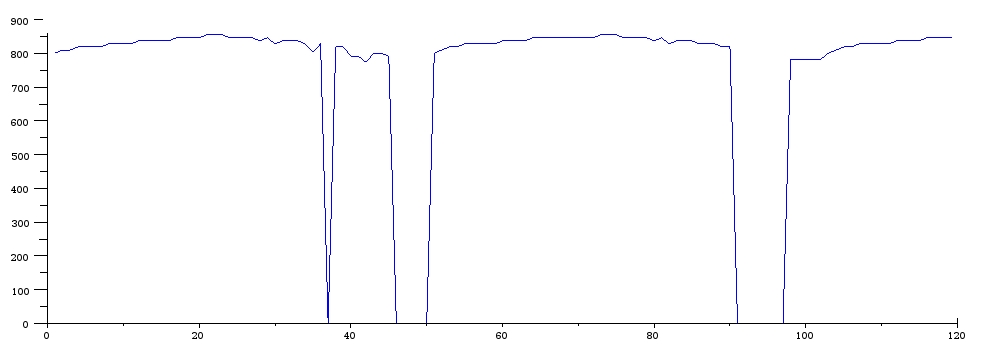}
     \vskip 0.1cm
     
     {Solutions with constraints (\ref{ctStretch1})}

      \caption{Illustration of an imposed modulation with stretch constraints on instance A3\_3\_210}\label{prodStretch}
   
\end{figure}

Adding CT6 constraints (\ref{defVarStretch}-\ref{ctStretch1}) or (\ref{defVarStretch}-\ref{ctStretch2}) 
has furthermore  few impact in the quality of the LP relaxation, which was already noticed in \cite{dupin2018dual}.
We analyze here whether it is insightful in searching primal solutions. 
Actually, decreasing profile phases are rarely activated in our model.
An explanation is that the upper bounds restrict the T2 production capacities.
The T2 units having the lowest marginal costs of production,
the  optimization avoids the over-costs of production with mainly maximal T2 productions.
Maximizing the T2 production available tends to avoid the decreasing profile phases,
preferring to have earlier outages to produce at the maximum power in the production cycles.
Optimal uses  of decreasing profile occur
when it is imposed by time windows constraints  to have longer  cycles than the duration to reach $\mathbf{Bo}_{i,k}$ fuel levels, which is  the situation met  in Figure \ref{prodStretch}.
In such cases, the  optimization tends  to produce at the upper bound of production
thanks to the difference in marginal costs between T1 and T2 units, which justifies the lighter formulation of CT6 constraints.


The over-costs to project the optimal (or best known) solutions computed thanks to the MILP of section 4.2
into a MILP with  stretch constraints is given for the small instances in Table \ref{stretchResult}.
Little over-costs are observed, around $0.3\%$ in average.
The full relaxation of CT6 constraints allows to generate solutions of very good quality after repairing of CT6.
Figure \ref{prodStretch} suggests that some local modifications around the CT6 situations can improve the previous projected solution.
 Table \ref{stretchResult}  gievs also the improvements of the projected solutions
 with only two iterations of local search with MILP neighborhoods $\N_{(0,2)}^{TW}$ using (\ref{ctStretch1}), which are tractable computations.
These very local modifications around the CT6 situations improve significantly the projected costs.

These conclusions justify the relaxation of CT6 constraints in a matheuristic approach, as developed in this paper, but also
in the approaches of \cite{lusby}.

\newpage

\section*{{Appendix B: Intermediate results}}


\begin{table}[ht]
     \caption{Comparison of termination time of B\&B to optimality:
     without and with the exact pre-processing of \cite{dupin2018dual},
     pre-processing and B\&B warmstart with the optimal (or best known) solution,
     and with penalisation costs to the baseline solution given by 5.1. These results were obtained using Cplex 12.5.}\label{tempsCalcul}
\centering
\begin{tabular}{|l|l|l|l|l|}
\hline
Instances&no PP&+ PP& +warmstart&penal\\
\hline
A1\_3\_120&0,12&0,12&0,04&0,13\\
A2\_3\_120&0,31&0,29&0,27&0,58\\
A3\_3\_120&0,23&0,21&0,21&0,14\\
A4\_3\_120&1,55&0,89&0,45&0,39\\
A5\_3\_120&127&82&70,86&15,3\\
\hline
 Total A&129,21&83,51&71,83&16,54\\
\hline
B6\_3\_120&32,3&17&13,8&4,98\\
B7\_3\_120&916,9&512&437,9&20,1\\
B8\_3\_120&$>$3600&1483&1461&45\\
B9\_3\_120&$>$3600&$>$3600&$>$3600&264\\
B10\_3\_120&60,8&52&43,8&5,4\\
\hline
X11\_3\_120&60,5&40,7&21,75&5,7\\
X12\_3\_120&111&69,4&43,6&17,2\\
X13\_3\_120&27,6&22&16,2&5,98\\
X14\_3\_120&98,8&56,1&53,2&16,3\\
X15\_3\_120&79,3&14&8,7&15,4\\
\hline
\end{tabular}
\end{table}

\begin{table}[ht]
      \centering
  \caption{Comparison of the gaps to the BKS for the POPMUSIC-VND and for the VND with single types of large neighborhoods}\label{PANvns2} 
\begin{tabular}{|l|l|l|l|l|l|}
\hline
Instances&Init&$\N_{(1,3)}^{TW}$&$\N_{k,k+1}^{cycles}$&$\N_{I}^{units}$&POPMUSIC-VND\\
&&&&$|\II|=8$&\\
\hline
B6&3,22\%&0,13\%&0,13\%&0,04\%&0,00\%\\
B7&3,68\%&0,35\%&0,4\%&0,02\%&0,00\%\\
B8&52,54\%&16,05\%&27,63\%&0,36\%&0,00\%\\
B9&27,46\%&7,54\%&12,12\%&0,05\%&0,00\%\\
B10&0,21\%&0,02\%&0,03\%&0,03\%&0,00\%\\
\hline
X11&0,45\%&0,22\%&0,26\%&0,09\%&0,00\%\\
X12&0,25\%&0,07\%&0,1\%&0,05\%&0,00\%\\
X13&7,37\%&0,93\%&0,28\%&0,07\%&0,00\%\\
X14&6,02\%&0,45\%&0,6\%&0,13\%&0,00\%\\
X15&0,12\%&0,06\%&0,08\%&0,05\%&0,00\%\\
\hline
Total &10,31\%&2,63\%&4,26\%&0,09\%&0,00\%\\
\hline
\end{tabular}

\end{table}

\begin{table}[ht]
      \centering
\caption{Result comparison for constructive primal math-heuristics}\label{tabPANconstr}
\begin{tabular}{ | l |l l|l|l l l|l|}
\hline
&BKS & Dual&Frontal&Simp&CMSA&R-R\&Fix &VND\\
\hline
B6&76966&0,13\%&0,00\%&3,22\%&3,8\%&0,34\%&0,00\%\\
B7&74234&0,52\%&0,02\%&3,68\%&0,75\%&0,25\%&0,00\%\\
B8&73240&7,95\%&NS&52,54\%&52,21\%&2,02\%&0,00\%\\
B9&72812&6,69\%&NS&27,46\%&28,95\%&1,15\%&0,00\%\\
B10&69501&0,07\%&0,02\%&0,21\%&0,2\%&0,21\%&0,00\%\\
\hline
X11&73018&0,37\%&0,00\%&0,45\%&0,53\%&0,38&0,00\%\\
X12&70604&0,20\%&0,00\%&0,25\%&0,24\%&0,31\%&0,00\%\\
X13&69231&1,09\%&0,04\%&7,37\%&9\%&0,71\%&0,00\%\\
X14&68395&0,89\%&0,02\%&6,02\%&1,21\%&0,54\%&0,00\%\\
X15&66029&0,10\%&0,06\%&0,12\%&0,12\%&0,38\%&0,00\%\\
\hline
Total&714031&0,00\%&NS&10,31\%&9,89\%&0,63\%&0,00\%\\
\hline
\end{tabular}
\end{table}

\begin{table}[ht]
      \centering
      
        \caption{Quality of local minimums considering one single type of neighborhoods in the VND of Algorithm 3}\label{vndLocalMin}
\begin{tabular}{|l|l|l|l|l|l|l|}
\hline
Instances&Init&$\N_{(0,1)}^{TW}$&$\N_{(1,2)}^{TW}$&$\N_{(0,3)}^{TW}$&$\N_{(1,3)}^{TW}$&$\N_{(0,5)}^{TW}$\\
\hline
B6&3,22\%&0,54\%&0,13\%&0,26\%&0,13\%&0,13\%\\
B7&3,68\%&1,14\%&0,41\%&0,47\%&0,35\%&0,19\%\\
B8&52,54\%&38,34\%&19,70\%&31,12\%&16,05\%&26,73\%\\
B9&27,46\%&15,91\%&7,54\%&12,35\%&7,54\%&8,89\%\\
B10&0,21\%&0,1\%&0,05\%&0,06\%&0,02\%&0,05\%\\
\hline
X11&0,45\%&0,31\%&0,10\%&0,26\%&0,22\%&0,27\%\\
X12&0,25\%&0,13\%&0,10\%&0,09\%&0,07\%&0,06\%\\
X13&7,37\%&2,23\%&1,30\%&1,47\%&0,93\%&1,01\%\\
X14&6,02\%&1,64\%&0,48\%&0,61\%&0,45\%&0,54\%\\
X15&0,12\%&0,08\%&0,06\%&0,07\%&0,06\%&0,06\%\\
\hline
Total&10,31\%&6,17\%&3,05\%&4,78\%&2,63\%&3,87\%\\
\hline
\end{tabular}
 
\vskip 0.5cm 
 
%
\begin{tabular}{|l|l|l l l|l l|}
\hline
&Init&$\N_{I}^{units}$&$\N_{I}^{units}$&$\N_{I}^{units}$&$\N_{k,k'}^{cycles}$&$\N_{k,k'}^{cycles}$\\
&&$|\II|=1$ &$|\II|=5$ &$|\II|=8$&$k=k'$&$k'-k=1$\\
\hline
B6&3,22\%&0,26\%&0,03\%&0,04\%&0,31\%&0,13\%\\
B7&3,68\%&0,34\%&0,06\%&0,02\%&0,55\%&0,40\%\\
B8&52,54\%&1,18\%&0,36\%&0,36\%&35,79\%&27,63\%\\
B9&27,46\%&1,31\%&0,05\%&0,05\%&14,48\%&12,12\%\\
B10&0,21\%&0,05\%&0,03\%&0,03\%&0,08\%&0,03\%\\
\hline
X11&0,45\%&0,16\%&0,09\%&0,09\%&0,30\%&0,26\%\\
X12&0,25\%&0,08\%&0,05\%&0,05\%&0,15\%&0,10\%\\
X13&7,37\%&0,51\%&0,07\%&0,07\%&2,43\%&0,28\%\\
X14&6,02\%&0,28\%&0,13\%&0,13\%&1,40\%&0,60\%\\
X15&0,12\%&0,09\%&0,05\%&0,05\%&0,10\%&0,08\%\\
\hline
Total&10,31\%&0,43\%&0,09\%&0,09\%&5,67\%&4,26\%\\
\hline
\end{tabular}

\end{table}

\begin{table}[ht]
\centering
\caption{Over-Costs of the outage solutions with the MILP of section 4.2
in the model adding CT6 constraints,
and local improvements with two iterations of VND with neighborhoods $\N_{(0,2)}^{TW}$.
}\label{stretchResult}
\begin{tabular}{|l|l|l|l|}
\hline
&BKS&Projected 4.2&+LS\\
\hline
A1\_3\_120&7612M&0,01\%&0,00\%\\
A2\_3\_120&7213M&0,01\%&0,00\%\\
A3\_3\_120&7638M&0,02\%&0,00\%\\
A4\_3\_120&5 122M&0,05\%&0,00\%\\
A5\_3\_120&5 852M&0,08\%&0,03\%\\
\hline
B6\_3\_120&31 950M&0,10\%&0,00\%\\
B7\_3\_120&30 364M&0,18\%&0,06\%\\
B8\_3\_120&29 438M&0,14\%&0,13\%\\
B9\_3\_120&29 239M&0,18\%&0,00\%\\
B10\_3\_120&27 957M&0,34\%&0,06\%\\
\hline
X11\_3\_120&30 199M&0,20\%&0,00\%\\
X12\_3\_120&29 669M&0,25\%&0,00\%\\
X13\_3\_120&29 148M&0,36\%&0,00\%\\
X14\_3\_120&28 438M&0,23\%&0,00\%\\
X15\_3\_120&27 722M&0,37\%&0,00\%\\
\hline
\end{tabular}

\end{table}

 \end{document}